\documentclass[sn-mathphys, Numbered, a4paper]{sn-jnl}

\usepackage{graphicx}%
\usepackage{multirow}%
\usepackage{amsmath,amssymb,amsfonts}%
\usepackage{amsthm}%
\usepackage{mathrsfs}%
\usepackage[title]{appendix}%
\usepackage{textcomp}%
\usepackage{manyfoot}%
\usepackage{booktabs}%
\usepackage{algorithm}%
\usepackage{algorithmicx}%
\usepackage{algpseudocode}%
\usepackage{listings}%
\usepackage{wrapfig}%
\usepackage{subfig}%
\usepackage[table,xcdraw]{xcolor}%
\usepackage{hyperref}%
\usepackage[utf8]{inputenc}%
\usepackage{cmap}
\usepackage{color}
\usepackage{lscape}
\usepackage{tabularray}

\usepackage{float}
\usepackage{tabularx, cellspace} %
\usepackage{geometry}
\begin{document}

\title[Article Title]{A case study of spatiotemporal forecasting techniques for weather forecasting}


\author*[1]{\fnm{Shakir Showkat} \sur{Sofi}}\email{Shakir.Sofi@skoltech.ru}

\author[1]{\fnm{Ivan} \sur{Oseledets}} 

\affil[1]{\orgname{CAIT, Skoltech.}, \orgaddress{\street{Bolshoy Bulvar}, \city{Skolkovo}, \postcode{121205}, \state{Moscow}, \country{Russia}}}

\abstract{The majority of real-world processes are spatiotemporal, and the data generated by them exhibits both spatial and temporal evolution. Weather is one of the most essential processes in this domain, and weather forecasting has become a crucial part of our daily routine. Weather data analysis is considered the most complex and challenging task. Although numerical weather prediction models are currently state-of-the-art, they are resource-intensive and time-consuming. Numerous studies have proposed time series-based models as a viable alternative to numerical forecasts. Recent research in the area of time series analysis indicates significant advancements, particularly regarding the use of state-space-based models (white box) and, more recently, the integration of machine learning and deep neural network-based models (black box). The most famous examples of such models are RNNs and transformers. These models have demonstrated remarkable results in the field of time-series analysis and have demonstrated effectiveness in modelling temporal correlations. It is crucial to capture both temporal and spatial correlations for a spatiotemporal process, as the values at nearby locations and time affect the values of a spatiotemporal process at a specific point. This self-contained paper explores various regional data-driven weather forecasting methods, i.e., forecasting over multiple latitude-longitude points (matrix-shaped spatial grid) to capture spatiotemporal correlations. The results showed that spatiotemporal prediction models reduced computational costs while improving accuracy. In particular, the proposed tensor train dynamic mode decomposition-based forecasting model has comparable accuracy to the state-of-the-art models without the need for training. We provide convincing numerical experiments to show that the proposed approach is practical.}


\maketitle

\section{Introduction}\label{sec1}

Time series forecasting has played a vital role in various real-world applications, including financial investments, traffic forecasting, supply chain management, epidemiology, and health monitoring. Weather forecasting is one of the essential applications of time-series analysis, which has been the most appealing research area in statistical physics, data assimilation, and data science. We all know that the sky can be clear one moment and cloudy, rainy, or completely different the very next. The chaotic nature of the weather makes weather forecasting extremely challenging. An accurate weather forecast is also crucial to many businesses and industries, such as agriculture and mining, tourism and food processing, sports, naval systems, airports, and PV projects (see e.g.,  \cite{irena, singh2018big, arroyo2011different, jones1986application, bilbao2002air}).  \par

Prior to the 1920s, weather forecasts were mainly reliant on weather charts. Meteorologists routinely prepared synoptic weather maps to forecast the average climate patterns. In 1922 \cite{nwp1922}, the author proposed a numerical model for weather prediction by formulating a set differential equation for climatic variables. The basic idea behind numerical weather prediction (NWP) is to solve the set of nonlinear PDEs (also known as primitive equations) over the different grid points in latitude and longitude space with known initial conditions. The solutions of these equations govern the atmospheric evolution \cite{nwp_eqn}. The first successful NWP with barotropic vorticity equations was presented in \cite{charney}. Since then, various methods have been proposed to improve and simplify forecasting by solving approximated equations. The approximate models have reduced computation costs and provided more control over model resolution (see, e.g., \cite{kalnay}). However, there are several issues with numerical weather prediction models, some related to the physics of meteorological variables and some technical. One fundamental issue with NWP is the uncertainty associated with initial conditions. Computational resources and computation time are two other significant issues with NWP models. Models with high spatial and temporal resolution require a lot of computing power, and current implementation methods take long model run times \cite{issuesNWP}. With these issues in mind and the enormous success of data-driven approaches, researchers are now focusing on data-driven weather forecasting methods. Many researchers have incorporated newly observed data and compared it to previous forecasts in order to update model states in such a way that the forecast trajectory remains close to true measurements; this approach is known as data assimilation, and it has been very successful in weather forecasting \cite{kalnay, Evensen}. Nowadays, several researchers employ data-driven approaches for preprocessing and postprocessing, while others have focused exclusively on data-driven techniques for weather forecasting. \par
Weather data is typically spatiotemporal, meaning it is collected across the space and time axes. Researchers have focused on capturing temporal correlations in state-space-based models for time series forecasting tasks. The research community has recently adopted AI-based models to capture spatiotemporal correlations; see, e.g., \cite{qin1, ConvLSTM1, rover, Bai2018AnEE, STConvS2S, Tran2018ACL, tan2023openstl, prednet, phydnet, simvp, tau} and the references therein. Research on incorporating spatiotemporal features into state-space-based models has been limited. This paper mainly focuses on regional data-driven weather forecasting, which involves forecasting over multiple latitude-longitude points to capture spatiotemporal correlations. The goal is to model the joint dynamics of data collected on a spatial grid (lat-lon) over time. This work aims to determine the extent to which existing interpretable models (state-space type white box models) can produce meaningful forecasts when spatiotemporal correlations are incorporated, with the following major contributions:
\begin{itemize} 
\item We set out to investigate the idea of including spatiotemporal correlations in different models.

\item We populate the idea by presenting two simple baseline spatiotemporal modelling approaches (sampling and clustering) that are based on dimensionality reduction, in which we identify important locations/subregions on the earth's surface using haversine distance-based clustering, train models on centred locations, and use these models to make forecasts at each location within their respective clusters. They significantly reduce computation compared to the naive strategy of training separate models on each latitude-longitude grid point by focusing on key locations and using cluster-based forecasting.

\item We adapt tensor-based dynamic mode decomposition (DMD) for spatiotemporal forecasting; the proposed methodology extends the classical DMD forecasting approach to learn and predict from 2D matrices or higher-order tensors rather than vectors. We evaluated and compared the proposed model on the real-world weather datasets to standard baselines.
\end{itemize}
\noindent Let us highlight a few essential notations used in this work before moving on to the next section.\\ 
\textbf{Notations \\}
\begin{tabularx}{\linewidth}{@{}SlX@{}}
$x, \mathbf{x}, \mathbf{X}$ & scalar, vector,  and matrix \\
$\underline{\mathbf{X}} \in \mathbb{R}^{n_1  \times\cdots\times n_d}$ & $d^{\small th}$order tensor  with $n_{l}$ denoting \\ & the size of the $l$th mode \\

$x_{i_{1}, i_{2}, \cdots i_{d}}= \underline{\mathbf{X}}_{i_{1}, \ldots, i_{d}}$ & A particular entry of $d^{\small th}$ order tensor \\ 
$ \mathbf{X^{\top}}, \mathbf{X^{-1}},  \mathbf{X^{\dagger}}$ & transpose, inverse, and pseudo-inverse \\ & of a matrix $\mathbf{X}$ \\
$\odot, \otimes$ & Khatri–Rao, Kronecker products\\
vec$(\mathbf{X})$, vec$(\underline{\mathbf{X}})$ & vectorization of $\mathbf{X}$ or $\underline{\mathbf{X}}$ \\

$\underline{\mathbf{A}}^{(l)},~ \underline{\mathbf{X}}^{(l)},~ \underline{\mathbf{Y}}^{(l)}$ & core tensors
 \end{tabularx}
\vspace{0.2cm}

\noindent The rest of the paper is organized as follows: Section~\ref{sec:review} describes the current state of the art in time series and spatiotemporal weather forecasting. Section~\ref{sec:STF} discusses various types of spatiotemporal weather forecasting models. In Section~\ref{section:TTDMD}, we provide a theoretical foundation for the tensor train based DMD (TT-DMD) and extend the classical DMD forecasting to TT-DMD based forecasting. The experimental results are presented in Section~\ref{sec:results}. The last section discusses the conclusion and future extensions.

\section{Related work}\label{sec:review}
 Data-driven forecasting models use past states to predict future states. Mathematically, we can write : 
\begin{equation}
\label{eqn:tsf}
\hat{\mathbf{x}}_{t+1}, \hat{\mathbf{x}}_{t+2}, \cdots, \hat{\mathbf{x}}_{t+h} = \mathbf{f}\left(\mathbf{x}_{t-T+1}, \cdots, \mathbf{x}_{t-1},  \mathbf{x}_{t} \right),
\end{equation} where $\mathbf{f}$ is the function that takes a set of past states and predicts $h$ future states. In the naive time series model, one of the simplest functions is the identity function, which sets the future state to the most recent state in the past. Another basic idea is to predict the future state by linearly regressing a set of previous states, known as autoregression. Similarly, the moving average model calculates the prediction by finding the arithmetic mean of the most recent past states. Jenkins and Box proposed a statistical model based on past sequences, the most famous being the auto-regressive integrated moving average (ARIMA). This model combined the features of both auto-regression and moving average \cite{box_jenkin}.
The ARIMA and the seasonal ARIMA were utilized for forecasting monthly rainfall and predicting monthly mean temperatures \cite{stoch_rain, ssaArima, sarima_nanjing}. More recently, machine learning-based models (learnable functions) have been used for time-series predictions. In \cite{svmxgb}, researchers used the support vector machine (SVM \cite{svmvapnik}) and the extreme gradient boosting (XGB) proposed in \cite{xgboost} to estimate global solar radiation. XGB and SVM performed very well in terms of accuracy and stability when compared with known empirical models \cite{chen1, scott1}. They claimed that XGB has comparable forecast accuracy while also being more stable. Researchers in \cite{mutlu, quzhang} presented an excellent way of using neural networks to model and forecast trends in time series data. Some researchers are also developing hybrid models, such as \cite{hybrid}, combining wavelet analysis with NN.  A hybrid method was proposed that decomposes the annual runoff time series into simple and interpretable component time series using singular spectrum analysis. ARIMA is then applied to each component time series. Finally, individual forecasts are combined using the correlation procedure. This hybrid model was found to outperform various other models by a significant margin \cite{ssaArima}. \par

The method incorporating local spatial effects in the classical autoregressive model, the so-called spatiotemporal autoregressive model (STAR), was proposed in \cite{SEMM, SEMMapp}. The results show that spatiotemporal correlations improve prediction accuracy significantly. However, these models require human-engineered spatiotemporal features, which are challenging to design \cite{STML} and therefore impractical for multidimensional problems. In order to tackle this issue, researchers have recently employed deep learning (DL) approaches. Specifically, they have combined convolutional neural networks (CNNs) to capture spatial features \cite{cnn1} with recurrent neural networks to account for temporal evolution \cite{en13246623}. The primary breakthrough model, a modified version of the classical LSTM, integrates convolution operations within the LSTM model to incorporate spatial features \cite{ConvLSTM1, STConvS2S, prednet, phydnet, simvp, tau}. More sophisticated models, such as the state-of-the-art Real-time Optical flow by Variational Methods for Echoes of Radar (ROVER) algorithm \cite{rover} and FourCastNet \cite{pathak2022fourcastnet}, which is based on adaptive Fourier Neural Operators, have been developed (see, e.g.,  \cite{deeptf, deeptemporal}). Recently, a novel model based on vision transformers was proposed that utilizes 3D deep networks with earth-specific priors to handle complex patterns in weather data effectively \cite{ViTforecast}. This model has demonstrated promising results in extreme weather forecasts and ensemble forecasts. Most of the DL models we described above consider spatiotemporal sequence forecasting as a video prediction problem, i.e., the model takes input as a set of $l$ frames and outputs the future $h$ frames, where $l$ is lookback, and $h$ is the forecast horizon. \par 
We restate our objective to employ state-space-type interpretable models, which ought to be innately capable of capturing spatiotemporal features, handling large datasets, and, most significantly, working well even if we do not have ample data for training, unlike DL models. Tensor decomposition-based state space models are one of the finest options. Tensor decomposition-based models have shown excellent performance in multiway learning and overcoming the curse of dimensionality. Interested readers are referred to the following excellent reviews on tensor decompositions:  \cite{kolda2009tensors, lieven2017tensors}, and the references therein.

\section{Spatiotemporal forecasting: baselines}\label{sec:STF}
This section presents a few ideas for modeling spatiotemporal correlations in classical models. These models will serve as baselines for comparison. There are two main approaches for conducting comparison (in general): one involves comparing different models, including the proposed model, on a specific task or dataset; the other involves comparing the proposed model to different tasks or datasets to assess its overall performance. In this study, we will employ the first approach.
\subsection{Sampling and clustering}
\begin{figure}[H]
 \centering
\includegraphics[width=0.97\linewidth, clip=true]{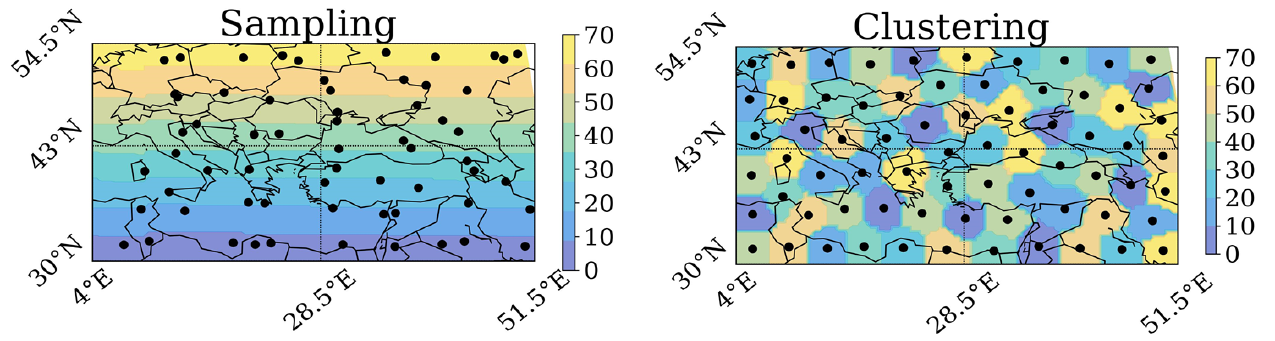}
\caption{On the left, we can see the locations sampled along the latitude direction, and the data from these locations is used for training. On the right, we can see the different clusters (regions), and the mean time-series data within a cluster is used for training.}
\label{fig:samp}
\end{figure}

Weather time series are collected at various latitude and longitude points, falling into the category of fixed spatial coordinates with measurements that change over time. For the classification of spatiotemporal datasets, please refer to \cite{Das_ST} and \cite{STML}. There are many ways to make predictions in this situation. One of the most straightforward ideas for forecasting time series at multiple locations is to use location-specific models to forecast at each location separately. This method has two major flaws: first, it is computationally expensive and requires a large amount of memory to save separate models, limiting its applicability to small areas; second, it ignores the effects of neighboring time series. In general, this approach is not feasible. Instead of training models at each location, one may select a subset of locations to train the models and then test/evaluate them at all locations, with each trained model being tested in its neighbourhood. The main issue now is how to sample these points in the region. One of the simplest methods is {\em naive sampling}, which randomly draws a desired number of points from latitude-longitude space. Naive sampling methods may be ineffective as they might overlook important regions. Although many factors influence climate, latitude direction is crucial when analyzing temperature, irradiance, etc., because different latitudes receive different amounts of solar energy \cite{lat-temp}, i.e., the temperature gradient changes more rapidly in the direction of latitude than in the direction of longitude. Therefore, a clever approach would be to sample more points in the latitude direction than in the longitude direction, as shown in Fig. \ref{fig:samp}.  Although the approach outlined above makes logical sense, it highly depends on the problem at hand. The nature of the underlying problem is often unknown to us, so we may not be able to determine the most important geospatial direction. Regarding more general problems, we prefer {\em clustering}, where the idea is to cluster geospatial locations based on some metric distance. In this case, the basic assumption is that the nearby locations have similar climate behavior, which generally holds. In the context of clustering spatiotemporal data, research presented in \cite{Dimred} suggests to use DBSCAN or to use haversine\footnote{\url{https://scikit-learn.org/stable/modules/generated/sklearn.metrics.pairwise.haversine_distances.html}} distance metric for the nearest-neighbor clustering. Finally, after clustering (sampling), the idea is to train the location-specific models (models respecting temporal correlations) on (central) time series indicated by the black dots in Fig. \ref{fig:samp} and then to perform prediction by evaluating the trained models at each location within its respective cluster (region).

\subsection{Snapshot LSTM}
\noindent In order to use the vector autoregression or LSTM model \cite{lstm_main}, \cite{gru}, \cite{sequenceLSTM}, \cite{encdecLSTM}, we vectorize a temperature field at each time instant, also called the "method of snapshots" in the fluid community. Consider a multivariate or multiple time series $\mathbf{Y} \in \mathbb{R}^{K \times T},$ where $K$ is the number of variables {\em (here in our case, K is the dimension of the vectorized temperature field or the number of longitude-latitude grid points)}, and $T$ is the total number of time-stamps. This reshaped dataset now consists of a single spatial dimension instead of two and a temporal dimension, i.e., each row of $ \mathbf{Y}$ represents a time series at a specific geospatial location, while each column corresponds to the temperature at all locations at a specific time instant (vectorized temperature field, referred to as a {\em snapshot}). The main problem with this method is that it removes inter-spatial relationships, which are often crucial. However, this approach enables us to utilize classical LSTM for evolving these snapshots over time as a multivariate time series.  \vspace{-3mm}
\subsection{Matrix autoregression}
\noindent In the above discussion of the snapshot approach, we have seen the main problem of vectorizing the maps (2D-temperature field), i.e., it removes the inter-spatial correlations. To tackle this problem, researchers in \cite{MAR} proposed the matrix autoregression,  the main idea of matrix autoregression (MAR) is to generalize the vector autoregression (VAR), an excellent attempt to incorporate spatial features. Consider a spatiotemporal tensor $\underline{\mathbf{X}} \in \mathbb{R}^{M \times N \times T}$ at each time $t$, we have a weather map as matrix $\mathbf{X}_{t}$ of size $M \times N$, where $M, N$ correspond to spatial locations/dimensions. The authors have formulated the first-order MAR model in a bilinear structure, as shown below:
\begin{equation}
 \vspace{-0.5mm}
\mathbf{X}_{t} = \mathbf{A}\mathbf{X}_{t-1}\mathbf{B^{\top}} +\mathcal{E}_{t},
\end{equation} where $\mathcal{E}_{t}$ is the error matrix at time $t$, $\mathbf{A}$ having size of $M \times M$, and $\mathbf{B}$ having size of $N \times N$. The authors claim that framing matrix values time series as above has two main benefits:
\begin{itemize}
\item This formulation offers substantial reduction of parameters as compared vectorization based. That is, MAR only needs $(M^2+N^2)$, while VAR for the same data will require $(MN)^2$ parameters. However, when problem size or regression order increases, coefficient estimations can become challenging. In such cases, alternative low-parametric solutions must be explored.
\item The coefficients $\mathbf{A}$ \& $\mathbf{B}$ are well interpretable depending upon problem at hand and an be easily estimated using gradient based methods.
\end{itemize}
\subsection{ConvLSTM}
\noindent Researchers proposed a model in \cite{ConvLSTM1} that enabled LSTM to include spatial features by using convolution operations within LSTM, i.e., convolution operator in input-to-state and state-to-state transition. Cell outputs, hidden states, input gate, output gate, and forget gate are 3D tensors with the last two dimensions corresponding to the spatial dimension, and they use spatial features inherited in design automatically. The key equations of ConvLSTM cell are shown below:
\begin{subequations}
\begin{align}
\mathbf{i}_{t} &= \sigma\left(\mathbf{W}_{i x} * \mathbf{X}_{t} + \mathbf{W}_{i h} * \mathbf{H}_{t-1} + \mathbf{W}_{i c} \circ \mathbf{C}_{t-1} + \mathbf{b}_{i}\right) \\
\mathbf{f}_{t} &= \sigma\left(\mathbf{W}_{f x} * \mathbf{X}_{t} + \mathbf{W}_{f h} * \mathbf{H}_{t-1} + \mathbf{W}_{f c} \circ \mathbf{C}_{t-1} + \mathbf{b}_{f}\right) \\
\mathbf{S}_{t} &= \tanh \left(\mathbf{W}_{c x} * \mathbf{X}_{t} + \mathbf{W}_{c h} * \mathbf{H}_{t-1} + \mathbf{b}_{c}\right)\\
\mathbf{C}_{t} &= \mathbf{f}_{t} \circ \mathbf{C}_{t-1} + \mathbf{i}_{t} \circ \mathbf{S}_{t} \\
\mathbf{o}_{t} &= \sigma\left(\mathbf{W}_{o x} * \mathbf{X}_{t} + \mathbf{W}_{o h} * \mathbf{H}_{t-1} + \mathbf{W}_{o c} \circ \mathbf{C}_{t} + \mathbf{b}_{o}\right) \\
\mathbf{H}_{t} &= \mathbf{o}_{t} \circ \tanh \left(\mathbf{C}_{t}\right)
\end{align}
\end{subequations}

\noindent where $\circ$ denotes Hadamard product and * denotes convolution operation.  ConvLSTM based encoder-decoder architecture has been quite successful in precipitation nowcasting \cite{ConvLSTM1}.
\section{Tensor based dynamic mode decomposition}
\label{section:TTDMD}
\noindent DMD is a dimensionality reduction algorithm that decomposes spatiotemporal data into different spatial/DMD modes, each with a temporal behavior defined by a single frequency and growth/decay rate (DMD eigenvalues) \cite{schmid, kutz}. This is similar to matrix factorization, in which a spatiotemporal matrix is decomposed into spatial and temporal factors. DMD has a wide range of applications; most notably, it is capable of forecasting multivariate time series.\par

\noindent Consider a  time-series data vectors $\left\{\mathbf{z}_{0}, \ldots, \mathbf{z}_{m}\right\}$, where each snapshots $\mathbf{z}_{k} \in \mathbb{R}^{n}.$  Assuming some (approximate linear) dynamical system generated the data, we can write:
\begin{equation}
\mathbf{z}_{k+1}=\mathbf{A}\mathbf{z}_{k},
\end{equation}
where matrix $\mathbf{A}$ is unknown,  $\mathbf{z}_{k}=\mathbf{z}(k \Delta t)$;  $\Delta t$ being fixed sampling rate. The main aim of DMD is to approximate $\mathbf{A}$, such that $\mathbf{Y} \approx \mathbf{AX},$ where $\mathbf{X}$ and  $\mathbf{Y}$ are snapshot matrices arranged as, $\mathbf{
X \triangleq\left[\begin{array}{lll}
z_{0} & \cdots & z_{m-1}
\end{array}\right], \quad Y \triangleq\left[\begin{array}{lll}
z_{1} & \cdots & z_{m}
\end{array}\right], }
$ we can find best-fit operator $\mathbf{A}=\mathbf{YX^{\dagger}}$ \cite{penrose}, but this operation is computationally expensive. DMD algorithm can approximate the dynamics of $\mathbf{A}$ by finding its eigenmodes and eigenvalues, as shown in the Algorithm~1.
\begin{figure}[ht]
  \centering
\includegraphics[width=0.94\textwidth]{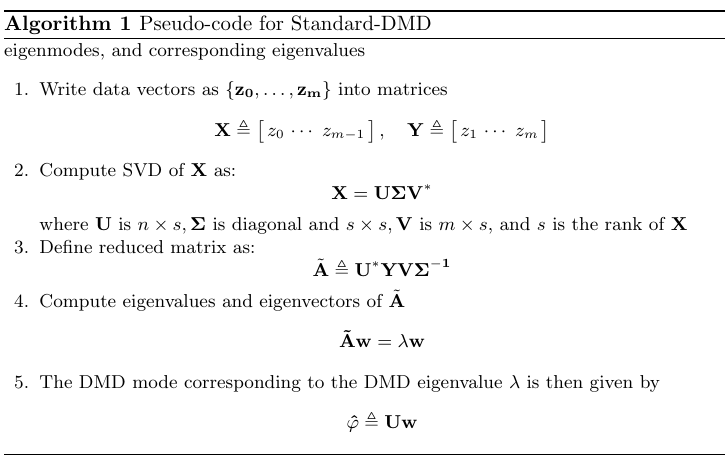}
  \label{alg:dmd1}
\end{figure}
\noindent After identifying the DMD modes and eigenvalues, one can easily reconstruct or forecast the snapshots using the following snapshot evolution equation:
\begin{equation}
\mathbf{z}_{k}=\boldsymbol{\Phi} \boldsymbol{\Lambda}^{k} \boldsymbol{\Phi}^{\dagger} \mathbf{z}_{0},
\end{equation} where $\boldsymbol{\Phi}$ is the matrix of whose columns are DMD modes ($ \hat{\varphi}'$s), and $\boldsymbol{\Lambda}$ is a diagonal matrix containing of eigenvectors of $\tilde{\mathbf{A}}.$  Note that the matrices $\mathbf{A}$
 and $\tilde{\mathbf{A}}$ share same eigenvalue spectrum.  This approach is reasonable, but it only applies to the 2D case, i.e., we have a snapshot vector at each time point. But what if each snapshot itself is a matrix or a multivariate tensor?  The multilinear variant of the DMD that utilizes tensor train decomposition was proposed in \cite{Klus}. This approach generalizes DMD to higher-order tensors. However, the paper did not demonstrate how to make predictions using this approach, which is the focus of our work.
 
\subsection{Tensor decompositions} The idea of tensor decomposition is to factorize the higher-order tensor into a set of smaller tensors that interact with each other with basic operations. The benefit of such factorization is that operations that we usually perform in data analysis can be performed very efficiently on smaller tensors, thus overcoming the curse of dimensionality. There are various tensor decompositions (a.k.a. formats), such as canonical polyadic decomposition (CPD \cite{hitchcock1927cpd, Harshman}), multilinear singular value  decomposition (MLSVD \cite{lieven2000mlsvd}) or tucker decomposition \cite{Tucker1966SomeMN}, hierarchical tucker format (hT- format \cite{gras2010hT}), and tensor train format (TT-format \cite{sirIO}). We will not go over these decompositions in this study. For details, please refer to \cite{kolda2009tensors, lieven2017tensors} and references therein. These decompositions offer alternative, efficient ways to perform linear and multilinear operations on large-scale data tensors in low-parametric form. The TT-format combines the best features of both CPD and MLSVD. It is stable like the MLSVD and overcomes the curse of dimensionality like the CPD. In recent years, the TT format has gained a lot of attention, and more than 10 Python packages offer fast implementations of linear and multilinear operations in the TT format. Thus, in this work, we use the TT format.

\subsubsection{TT format}\label{subsec:ttformat}

Let $\underline{\mathbf{A}} \in \mathbb{R}^{n_1  \times\cdots\times n_d}$ be the $d$th order tensor with $n_{l}$ denoting the size of the $l$th mode. In TT format  $\underline{\mathbf{ A}}$ is factored into $d$ core tensors $\underline{\mathbf{A}}^{(l)} \in$ $\mathbb{R}^{{r}_{l-1} \times n_{l} \times r_{l}}.$  The tuple of minimal integers $(r_0, r_1, \ldots, r_{d})$ for which equality in Equation \ref{eq:tt1} and Equation \ref{eq:tt2} holds is the TT-rank of $\underline{\mathbf{ A}}$. Entry-wise we can write: \begin{equation}\label{eq:tt1}
\underline{\mathbf{A}}_{i_{1}, \ldots, i_{d}}=\sum_{k_{0}=1}^{r_{0}} \ldots \sum_{k_{d}=1}^{r_d} \underline{\mathbf{A}}_{k_{0}, i_{1}, k_{1}}^{(1)} \cdots \cdot \underline{\mathbf{A}}_{k_{d-1}, i_{d}, k_{d}}^{(d)},\end{equation} where indices are written in subscripts. Furthermore, for two vectors $\boldsymbol{v} \in \mathbb{R}^{n_{1}}$ and $w \in \mathbb{R}^{n_{2}}$, the tensor product $\boldsymbol{v} \otimes \boldsymbol{w} \in \mathbb{R}^{n_{1} \times n_{2}}$ is given by $(\boldsymbol{v} \otimes \boldsymbol{w})_{i, j}=\left(\boldsymbol{v} \cdot \boldsymbol{w}^{\top}\right)_{i, j}=$
$v_{i} \cdot w_{j} .$ In this notation a tensor $\underline{\mathbf{A}}$ can be written as: \begin{equation}\label{eq:tt2}
\underline{\mathbf{A}}=\sum_{k_{0}=1}^{r_{0}} \ldots \sum_{k_{d}=1}^{r_{d}} \underline{\mathbf{A}}_{k_{0},:, k_{1}}^{(1)} \otimes \cdots \otimes\underline{\mathbf{A}}_{k_{d-1},:,k_{d}}^{(d)}. \end{equation} 
Additionally, we introduce the notation for matricizations (unfoldings) and vectorizations. For the two ordered disjoint subsets $N^{\prime}=\left\{n_{1}, \ldots, n_{l}\right\}$ and $N^{\prime \prime}=$ $\left\{n_{l+1}, \ldots, n_{d}\right\}$ of $N=\left\{n_{1}, \ldots, n_{d}\right\}$, the matricization of $\underline{\mathbf{A}}$ with respect to $N^{\prime}$ and $N^{\prime \prime}$ is denoted by $\underline{\mathbf{A}} \mid \begin{array}{l}N^{\prime \prime} \\ N^{\prime}\end{array} \in \mathbb{R}^{\left(n_{1} \cdots \cdot n_{l}\right) \times\left(n_{l+1} \cdots \cdot n_{d}\right)}$. That is, the first $l$ indices of the tensor enumerate rows of a reshaped matrix, and the remaining indices enumerate the columns. In literature, it is also called $l$-mode matrix unfolding. The special case where $N^{\prime}=N$ and $N^{\prime \prime}=\varnothing$, is  vectorization of $\underline{\mathbf{A}},$ denoted by vec$(\underline{\mathbf{A}}) \in \mathbb{R}^{n_{1} \cdots \cdot \cdot n_{d}}.$

\subsubsection{Tensor train-DMD}
\begin{figure}[ht]
  \centering
\includegraphics[width=0.5\textwidth]{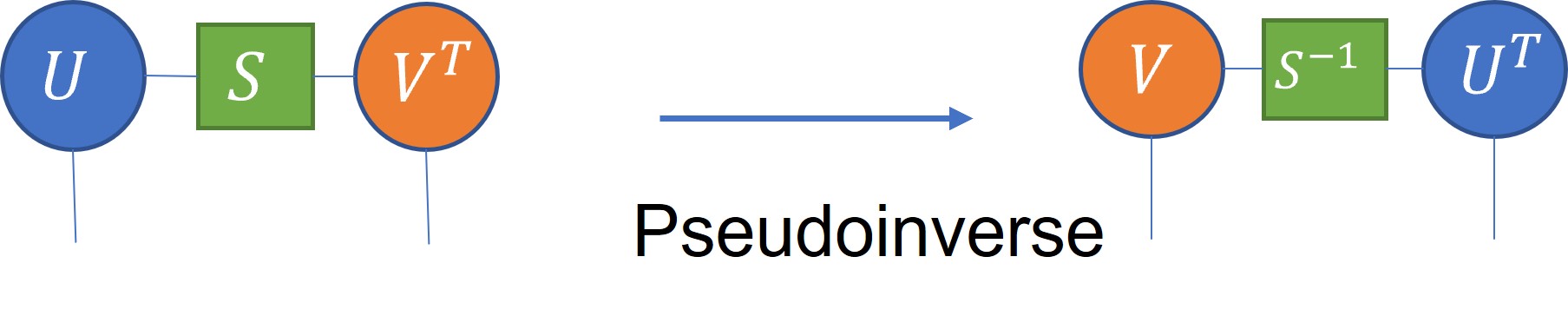}
    \caption{Pseudoinverse of a matrix using the SVD.}
  \label{fig:mpinv}
  \vspace{-5mm}
\end{figure}

\begin{figure}[ht]
  \centering
\includegraphics[width=0.94\textwidth]{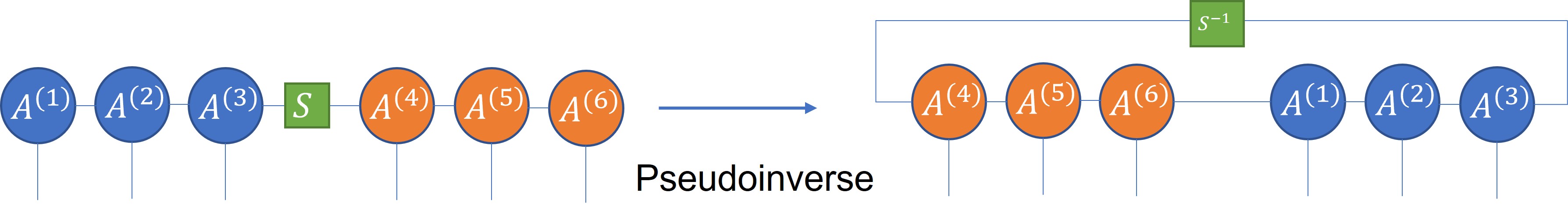}
\caption {Pseudoinverse of $3$-mode unfolding matrix of a sixth-order tensor in the TT format.}
  \label{fig:tpinv}
  \vspace{-5mm}
\end{figure}

In the previous section, we saw that the foremost expensive step in the DMD algorithm is computing the pseudoinverse. In the following section, we efficiently compute the pseudoinverse in TT format.  
The TT format is an SVD-like decomposition of higher-order tensors.  In the second-order case, we compute the pseudoinverse using SVD by reordering the orthonormal basis matrices and taking the inverse of the diagonal singular value matrix, as shown in Fig. \ref{fig:mpinv}. A similar approach that generalizes this idea was proposed in the TT format \cite{sirIOpinv, Klus}.
Interestingly, it was found that the TT format with the orthogonal TT cores around the unfolding mode possesses similar properties, i.e., we can compute pseudoinverse by reordering the left and right orthogonal TT cores. Fig.~\ref{fig:tpinv} shows the pseudoinverse of the unfolding matrix of a sixth-order tensor in TT-format with respect $N^{\prime}=\left\{n_{1},n_{2}, n_{3}\right\}$ and $N^{\prime \prime}=$ $\left\{n_{4}, n_{5}, n_{6}\right\}$.
For the tensor-based DMD, consider $m$ snapshots of $d$-dimensional tensor-trains $\underline{\mathbf{X}}, \underline{\mathbf{Y}} \in \mathbb{R}^{n_{1} \times \cdots \times n_{d} \times m}$,
where $\underline{\mathbf{X}}_{:, \ldots, :, i+1} \in \mathbb{R}^{n_{1} \times \cdots \times n_{d}}$ for $i=0, \ldots, m-1$ and $\underline{\mathbf{Y}}_{:, \ldots, :, i}$ for
$i=1, \ldots, m .$ Let the tuples of integers $(r_{0}, \ldots, r_{d+1})$ and $(s_{0}, \ldots, s_{d+1})$ be the TT-ranks of $\underline{\mathbf{X}}$ and $\underline{\mathbf{Y}}$, respectively. Now, let $\mathbf{X}, \mathbf{Y} \in \mathbb{R}^{n_{1} \cdots \cdot n_{d} \times m}$ be the matricizations of $\underline{\mathbf{X}}$ and $\underline{\mathbf{Y}}$ along mode $m$. Each column in the respective matricizations corresponds to a snapshot. Let us decompose $\underline{\mathbf{X}}$ in TT format to compute the TT-DMD in low parametric representation efficiently. Let the SVD of $\mathbf{X}=\mathbf{M} \boldsymbol{\Sigma} \mathbf{N}.$ We can find $\mathbf{M}$ and $\mathbf{N}$ in TT format as:
$$\mathbf{M}=\left(\sum_{k_{0}=1}^{r_{0}} \cdots \sum_{k_{d-1}=1}^{r_{d-1}} \underline{\mathbf{X}}_{k_{0}, :, k_{1}}^{(1)} \otimes \cdots \otimes \underline{\mathbf{X}}_{k_{d-1}, :,:}^{(d)}\right) \Biggm| \begin{array}{l}r_{d} \\ n_{1}, \ldots, n_{d}\end{array}.$$
$\mathbf{N}=\left.\left(\underline{\mathbf{X}}_{:,:, k_{d+1}}^{(d+1)}\right)\right|_{r_{d}} ^{m}$, with $\boldsymbol{\Sigma}$ being the diagonal matrix with singular values on its diagonal elements, which are  computed by performing the SVD of
$\underline{\mathbf{X}}^{(d)} \mid \begin{array}{l}r_{d} \\ r_{d-1}, n_{d}\end{array}$ Note that $\mathbf{N}$ is equivalent to the last core $\underline{\mathbf{X}}^{(d+1)}$, which is
a matrix because $r_{d+1}=1$. This procedure is similar to the SVD for matrices, but in the TT format, we work on smaller cores instead of full matricization. $\mathbf{M} \in \mathbb{R}^{n_{1} \cdots n_{d} \times r_{d}}$ is computed by left-orthogonalization of first $d$ cores of $\underline{\mathbf{X}}$ (see, left and right orthogonalization of TT cores \cite{sirIO}). Next, we assume a linear relationship between the pairs of data vectors as in the matrix case, i.e.
\begin{equation}
\mathbf{Y=A X},
\end{equation}
with $\mathbf{A} \in \mathbb{R}^{n_{1} \cdots n_{d} \times n_{1} \cdots n_{d}} ,$ where  $\mathbf{A = Y \cdot X^{\dagger}}$. The pseudoinverse $\mathbf{X^{\dagger}}$ can be written as:
$$
\mathbf{X^{\dagger}}=\left(\underline{\mathbf{X}} \Biggm| \begin{array}{l}
m \\
n_{1}, \ldots, n_{d}
\end{array}\right)^{\dagger}=\mathbf{N}^{\top} \boldsymbol{\Sigma}^{-1} \mathbf{M}^{\top}.
$$
Similarly, the unfolding $\underline{\mathbf{Y}}$ can also be written in the TT format:
\begin{equation*}
\begin{split}
& \mathbf{Y}=\left.\underline{\mathbf{Y}}\right|_{n_{1}, \ldots, n_{d}} ^{m}. \\
& 
=\left.\left(\sum_{l_{0}=1}^{s_{0}} \cdots \sum_{l_{d-1}=1}^{s_{d-1}} \underline{\mathbf{Y}}_{l_{0}, :, l_{1}}^{(1)} \otimes \ldots \otimes \underline{\mathbf{Y}}_{l_{d-1},:,:}^{(d)}\right)\right|_{n_1, \cdots, n_d}^{s_{d}}\left.\cdot \underline{\mathbf{Y}}^{(d+1)}\right|_{s_{d}}^{m}=\mathbf{P Q}.
\end{split}
\end{equation*}

\noindent Using the low-parametric representation of $\mathbf{X}^{\dagger}$ and $\mathbf{Y}$ mentioned above, we can extend the matrix algorithm to higher-order cases. We can rewriting the expression for the reduced matrix $\tilde{\mathbf{A}}$ in decomposed form as:
\begin{equation}
\tilde{\mathbf{A}}=\mathbf{M}^{\top} \cdot \mathbf{P Q \cdot N}^{\top} \boldsymbol{\Sigma}^{-1}.
\end{equation}
For efficient computation, we split the above expression into different parts. First, consider $\mathbf{M^{\top} \cdot P}.$ Each entry at position $(i, j)$ can be expressed in TT format as:  
\begin{equation*}
\begin{split}
& \left(\mathbf{M}^{\top} \cdot \mathbf{P}\right)_{i, j}\\ & = \sum_{k_{0}=1}^{r_{0}} \cdots \sum_{k_{d-1}=1}^{r_{d-1}} \sum_{l_{0}=1}^{s_{0}} \cdots \sum_{l_{d-1}=1}^{s_{d-1}}\left(\underline{\mathbf{X}}_{k_{0}, : , k_{1}}^{(1)}\right)^{\top} \underline{\mathbf{Y}}_{l_{0}, : , l_{1}}^{(1)} \ldots\left(\underline{\mathbf{X}}_{k_{d-1}, :, i}^{(d)}\right)^{\top} \underline{\mathbf{Y}}_{l_{d-1}, :, j}^{(d)}.
\end{split}
\end{equation*}

Using the following property in TT format: $$\operatorname{vec}(\underline{\mathbf{X}})^{\top} \cdot \operatorname{vec}(\underline{\mathbf{Y}})=\Pi_{l=1}^{d}\left(\underline{\mathbf{X}}^{(l)}\right)^{\top} \cdot \underline{\mathbf{X}}^{(l)} .$$ We compute  $\mathbf{M}^{\top} \cdot \mathbf{P}$ in TT format very efficiently since we only need to reshape some TT cores, and use contractions (see, Algorithm 4 \cite{sirIO}). Assuming TT ranks of $\underline{\mathbf{X}}$ and $\underline{\mathbf{Y}}$ are small with respect to the whole state-space. The unfolding ranks $r_{d}$ and $s_{d}$ are both bounded by  $m$. We can write $\mathbf{Q} \cdot \mathbf{N}^{\dagger}=\left.\underline{\mathbf{Y}}^{(d+1)}\right |_{s_{d}} ^{m} \cdot\left(\underline{\mathbf{X}}^{(d+1)} \Biggm| \begin{array}{l}m \\ r_{d}\end{array}\right)^{\dagger}$.\\
Considering the exact DMD method outlined as Algorithm-2 in \cite{on_dmd}, we can compute the DMD modes of $\tilde{\mathbf{A}}$. If $\lambda_{1}, \ldots, \lambda_{p}$ are the eigenvalues of  $\tilde{\mathbf{A}}$ and the  corresponding eigenvectors are $\boldsymbol{w}_{1}, \ldots, \boldsymbol{w}_{p} \in \mathbb{R}^{r_{d+1}}$, then the vectorized DMD modes, i.e.,  $\mathbf{\varphi}_{1}, \ldots, \mathbf{\varphi}_{p} \in \mathbb{R}^{n_{1} \cdots \cdot \cdot n_{d}}$ of $\mathbf{A}$ can be expressed as:   \begin{equation}\mathbf{\varphi}_{j}= \left(1 / \lambda_{j}\right) \cdot \mathbf{P} \mathbf{Q} \cdot \mathbf{N}^{\dagger} \boldsymbol{\Sigma}^{-1} \cdot \mathbf{w}_{j},\end{equation} for $j=1, \ldots, p$. All these modes are present in a single tensor, i.e., $\underline{\boldsymbol{\Phi}} \in \mathbb{R}^{n_{1} \times \cdots \times n_{d} \times p}$ which  is given by:
\begin{equation}\begin{aligned}\underline{\boldsymbol{\Phi}}=\sum_{l_{0}=1}^{s_{0}} .. \sum_{l_{d}=1}^{s_{d}} \underline{\mathbf{Y}}_{l_{0},:, l_{1}}^{(1)} \otimes \cdots  \otimes  \underline{\mathbf{Y}}_{l_{d-1},:, l_{d}}^{(d)} \otimes\left(\mathbf{Q} \cdot \mathbf{N}^{\dagger} \boldsymbol{\Sigma}^{-1} \cdot \mathbf{W} \cdot \boldsymbol{\Lambda}^{-1}\right)_{l_{d},:} ,\end{aligned}\end{equation}\\
where  $\operatorname{vec}\left(\underline{\boldsymbol{\Phi}}_{:, \ldots,:, j}\right)=\mathbf{\varphi}_{j}$ and $\boldsymbol{\Lambda}$ is a diagonal matrix of eigenvalues. The summary of the whole algorithm method is outlined as Algorithm~2:
\begin{figure}[ht]
  \centering
\includegraphics[width=0.99\textwidth]{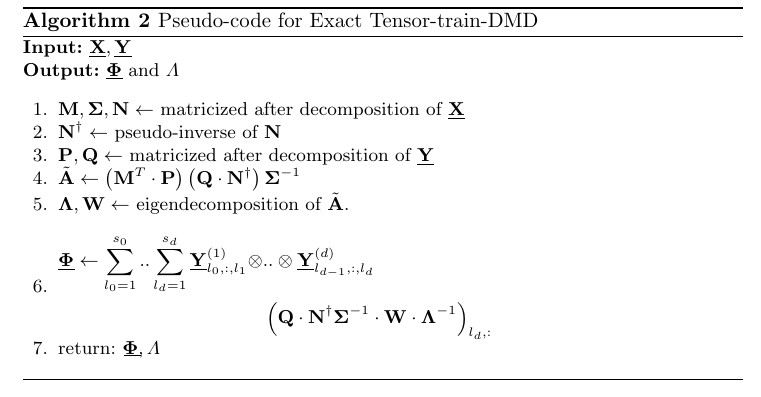}
  \label{alg:tdmd}
\end{figure}
\subsubsection{Forecasting with TT-DMD}
\noindent DMD-based forecasting has received a lot of attention and has been investigated in several articles. In \cite{sashidhar2022bagging}, researchers have proposed the optimized DMD algorithm with statistical bagging, and it was found that it can make stable forecasts. Researchers in \cite{dylewsky2022sfdmd} have used a DMD-based forecasting model for power load predictions and have shown how the proposed model learns a stochastic Gaussian process regression (see, e.g.,  \cite{yue2022pdmd, liew2022streaming, cheng2022real, mansouri2023weather}).  Multivariate forecasting with DMD using a block Hankel matrix was proposed in \cite{mch_DMD}. The main idea of DMD forecasting approaches is to learn an (approximate) linear model in the time-delay basis representation, i.e., it is assumed that the snapshot data is generated by a high-dimensional linear dynamic model or by a nonlinear model (which can be approximated with a high-dimensional linear model). Under this assumption, we learn this linear operator from the time-shifted copies of input data. After learning this linear dynamic operator, we evolve the states by feeding the most recent past state to generate forward point forecasts so it evolves as a linear dynamical system. However, these models cannot forecast matrices or higher-order tensors since they learn vector autoregression. For example, we cannot use these models directly to predict the temperature distribution over a spatial grid (a matrix where spatial coordinates enumerate the rows and columns); instead, a generalization of the forecasting approach that can inherently learn a multidimensional linear dynamic operator that can be used to predict the temperature distribution over spatial grids should be developed. This is the main motivation for this work. Based on the preceding section, we develop the forecasting method, which predicts 2D weather maps (temperature distributions over spatial grids) using a learned multilinear dynamic operator $(\underline{\boldsymbol{\Phi}})$. \par

\noindent Suppose that we have learned the low-rank multilinear operator $\underline{\boldsymbol{\Phi}} \in \mathbb{R}^{n_{1} \times \cdots \times n_{d} \times {p}}$ in TT-format from the multidimensional data tensor $\underline{\mathbf{X}}$ and its time-delayed shifted data tensor $\underline{\mathbf{Y}}$, as shown in Algorithm~2. Note that the dimensions of the data tensor $\underline{\mathbf{X}}$ is $d+1$, where the first $d$ indices correspond to different variables, and the last index corresponds to the time dimension. For example, if our dataset consists of 2D field distributions of temperature $(t)$ and pressure $(q)$ variables over the spatial 2D grid collected $(lat\times lon)$ for $(T)$ time points, the size of the data tensor $\underline{\mathbf{X}}$ will be $t \times q \times lat \times lon \times T$, meaning that it generalizes to any number of climate variables or side information (we only need extra channels). Thus, the TT format gave us everything we needed to generalize the DMD forecasting algorithm to higher-order tensors. After learning multilinear operator $\underline{\boldsymbol{\Phi}},$ we can find the coefficients vector $(b)$ by contracting the data tensor at initial time point $\underline{\mathbf{X}_{0}}$ with the pseudoinverse of multilinear operator $\underline{\boldsymbol{\Phi}}^{\dagger}$, this operation can be done very efficiently in tensor-train format by using the TT cores of $\underline{\boldsymbol{\Phi}}^{\dagger}$ and $\underline{\mathbf{X}_{0}}.$ Finally, to reconstruct and predict the data tensor, we need to contract the operator $\underline{\boldsymbol{\Phi}}$ with the time evolution of coefficient vectors based on the eigenvalues (Vandermond type evolution). Algorithm 3 shows the pseudo-code for the simplest case where only temperature distribution is observed over $M \times N$ spatial grid. In the numerical experiment section, we provide further details for its usage.
\begin{figure}[htb]
  \centering
\includegraphics[width=0.99\textwidth]{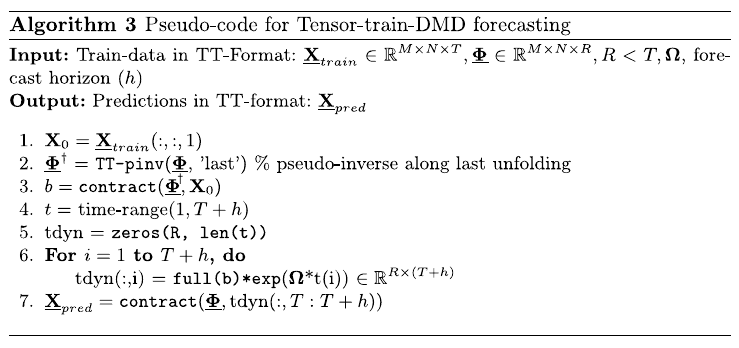}
  \label{alg:tdmdpred}
\end{figure}
\noindent where $\Omega$ is a vector of the logarithm of eigenvalues. The sample code is available\footnote{ \url{https://github.com/ShakirSofi/TensorizingDMD}} 
The algorithm shown above is designed to work with the original weather map dimensions and can reconstruct and predict matrices directly.

\section{Experiments and discussion of the results}
\label{sec:results}
\subsection{Dataset description}
\noindent The NASA Power database was used\footnote{The NASA Langley Research Center (LaRC) POWER Project funded through the NASA Earth Science/Applied Science Program: \url{https://power.larc.nasa.gov}}. This is a global database of regular weather data, primarily for agricultural purposes. The spatial resolution is 0.5°x 0.5°, with a temporal resolution of one day. We used the maximum temperature time series (TMAX in degrees Celsius). The selected subset of the data is formatted as a time series collected across latitude-longitude coordinate points between the 30-Oct-2015 and 15-November-2020 time range, containing approximately 1844 timestamps at each coordinate point (i.e., the length of each time series). We chose specific geographic locations, with a range of lat $ \times$ lon: [30.0° to 54.5°] $ \times$ [4.0° to 51.5°]. Around 4700 latitude-longitude points ($50 \times 94$) are contained within this latitude-longitude range. This latitude-longitude range encompasses parts of Europe and Central Asia. 
\subsection{Evaluation criteria}
\noindent To measure the model performance, we can use different kinds of metrics. These metrics estimate the forecasting precision. Generally, these metrics measure the error between actual and predicted trajectories. Depending upon the need, we may use scale-dependent metrics, i.e., on the same scale on which the data itself is, for example, root mean square error (RMSE), mean absolute error (MAE), or scale-independent, which are more general, for example, absolute percentage error (MAPE, SMAPE). Here, we have used the following metrics to measure the performance, see \cite{perfmet}.\par
\begin{enumerate}
\item Root mean square error:
$$
\mathbf{R M S E}=\sqrt{\frac{1}{n} \sum_{i=1}^{n}\left(y_{i}^{\mathrm{obs}}-y_{i}^{\mathrm{pred}}\right)^{2}}.
$$
\item Mean absolute error:
$$
\mathbf{M A E}=\frac{\sum_{i=1}^{n}\left|y_{i}^{\mathrm{obs}}-y_{i}^{\mathrm{pred}}\right|}{n}.
$$
\item Symmetric mean absolute percentage error:
$$\mathbf{SMAPE}=\frac{100 \%}{n} \sum_{t=1}^{n} \frac{\left|y^{\mathrm{pred}}_{i}-y^{\mathrm{obs}}_{i}\right|}{\left(\left|y^{\mathrm{obs }}_{i}\right|+\left|y^{\mathrm{pred}}_{i}\right|\right) / 2}.$$
where $y_{i}^{\mathrm{obs}}$ and $y_{i}^{\mathrm{pred}}$ are the real and predicted data observations, respectively.
\end{enumerate}
\subsection{Numerical Experiments}
\noindent 
The experiments are performed to predict the temperature time series across all longitude-latitude coordinate points ([30.0° to 54.5°] $ \times$ [4.0° to 51.5°]),  in the following way: First, we make the short-term predictions (i.e., the next seven days) using all the spatiotemporal forecasting approaches described in section \ref{sec:STF}, and then, based on a performance comparison, we select models which are computationally less expensive to make long-term predictions (250 steps ahead) and compare their performance to ensure their ability to predict temperature fields for multiple time steps ahead.\par
\noindent \underline{Train-test split:} In the short-term prediction case, we choose a subset of data from \texttt{2015-10-30} to \texttt{2019-12-07}  for training and \texttt{2019-12-08} to \texttt{2019-12-14} for testing.  For the long-term prediction experiments, the training data is the same as that of the short-term case, but the test data ranges from \texttt{2019-12-08} to \texttt{2020-08-13}.

\subsubsection{Short-term predictions:}
\noindent \underline{Training procedure:}~ For sampling and clustering-based approaches, we sampled 70 locations as described in the section \ref{sec:STF},  also shown in Fig. \ref{fig:samp}.  We trained a 2-layer LSTM model at each of the 70 selected locations. We trained each LSTM model on centered  time series (time series calculated by averaging all time series within a given cluster/sub-region), with a time-lag {\em (a.k.a., lookback)} = 5 days, output sequence length = 7,  with 500 epochs, and a learning rate = 2.3e-4.  We optimize these models over MSE loss, i.e., \[L(\mathbf{y}, \hat{\mathbf{y}}) = \frac{1}{N} \sum_{i=0}^{N}(y_i - {\hat{y}}_i)^2,\] where $\mathbf{y},~ \hat{\mathbf{y}}$, are the actual and predicted values, respectively. The training procedure of Snapshot LSTM is identical to the above methods, except that in this model, the input data is the set of vectorized temperature fields (a set of vectors where, in each vector, we have temperature values collected across all locations). \par

\noindent ConvLSTM data preparation is similar to the LSTM data preparation, except that frames are used instead of vectors; data instances consist of twelve frames, five for input and seven for prediction. Similarly to what was proposed in the original work \cite{ConvLSTM1}, the four-layer ConvLSTM model was trained with each layer containing a hidden size of 32 and a $3\times 3$  kernel size. This model was also optimized on MSE loss with Adam optimizer $(lr=5e-3$, decay rate$ = 0.9)$ for 50 epochs. ConvLSTM, TT-DMD, and matrix autoregression (MAR) models work with the original dimensions of data and are capable of directly predicting a two-dimensional $(50 \times 94)$ temperature field for the next seven time steps, i.e, $\underline{\mathbf{X}}_{\text{predition}} \in \mathbb{R}^{50 \times 94 \times \text{steps}}$.  The matrix auto-regression (MAR) of order one was chosen as the baseline, and the MAR coefficient was learned using the ALS algorithm with $500$ iterations. We prepared shifted tensors $(\underline{\mathbf{X}}, \underline{\mathbf{Y}}),$ for TT-DMD and used Algorithm 2 from section \ref{section:TTDMD} to find TT-DMD modes ($\underline{\boldsymbol{\Phi}}$) and eigenvalues with the tensor train rank, $\text{rank}_{TT}  = [1,  25,  70,  1]$ (optimal by grid search). Note that if the data is noisy,  prior smoothening with low-rank tensor decompositions is very useful to reduce the $\text{rank}_{TT}$ (see, e.g.,  \cite{pmlr-v70-imaizumi17a}, \cite{yakota2016smooth}). \par

\begin{figure}[ht]
  \centering
\includegraphics[width=1\textwidth,clip, trim={0 8.5cm 0 0cm}]{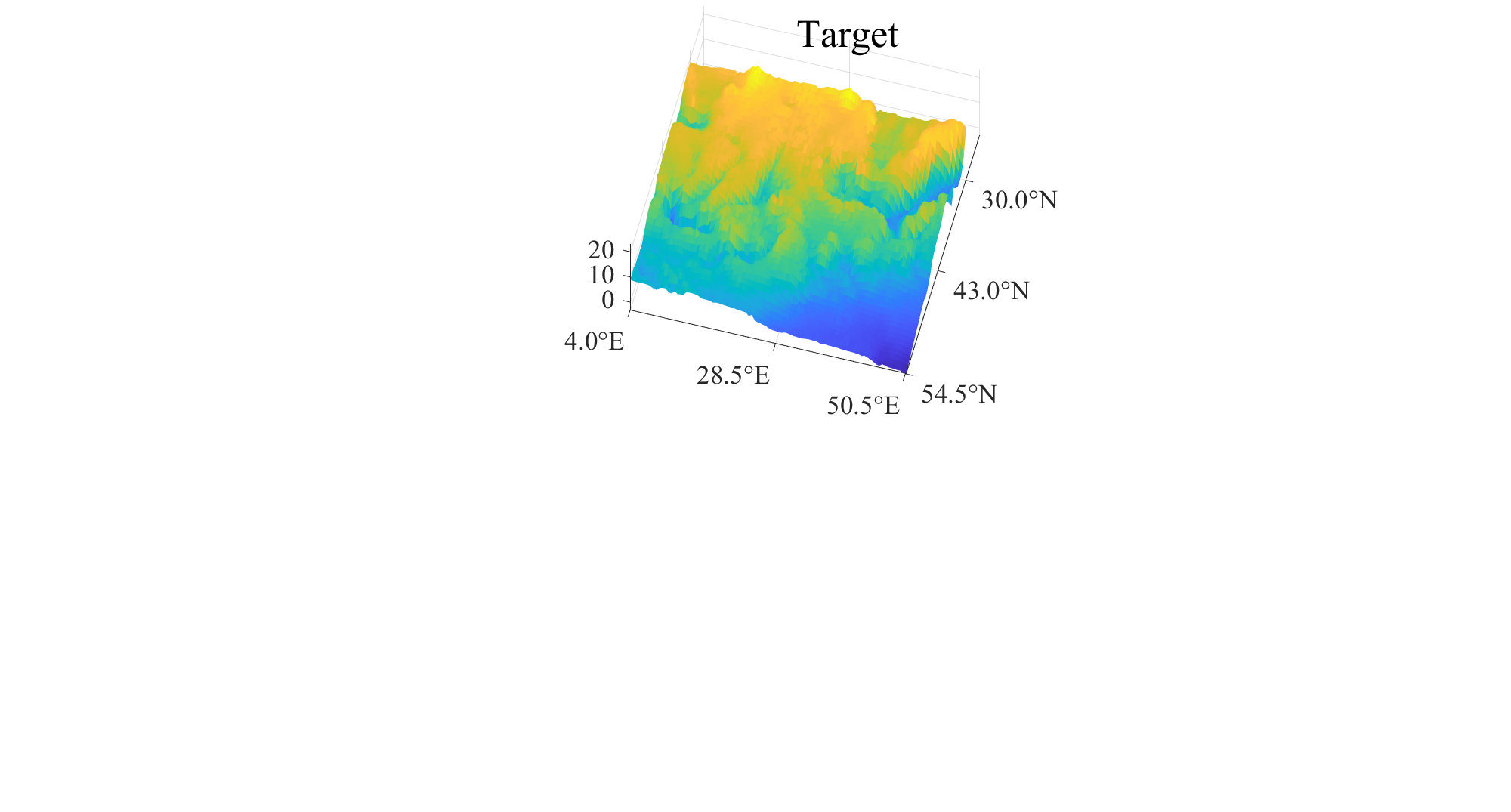}
    \caption{ Target temperature distribution \texttt{2020-08-13}}
    \label{fig:tarpred}
  \end{figure}

\begin{figure}[ht]
\vspace{-6mm}
  \centering
\includegraphics[width=1\textwidth]{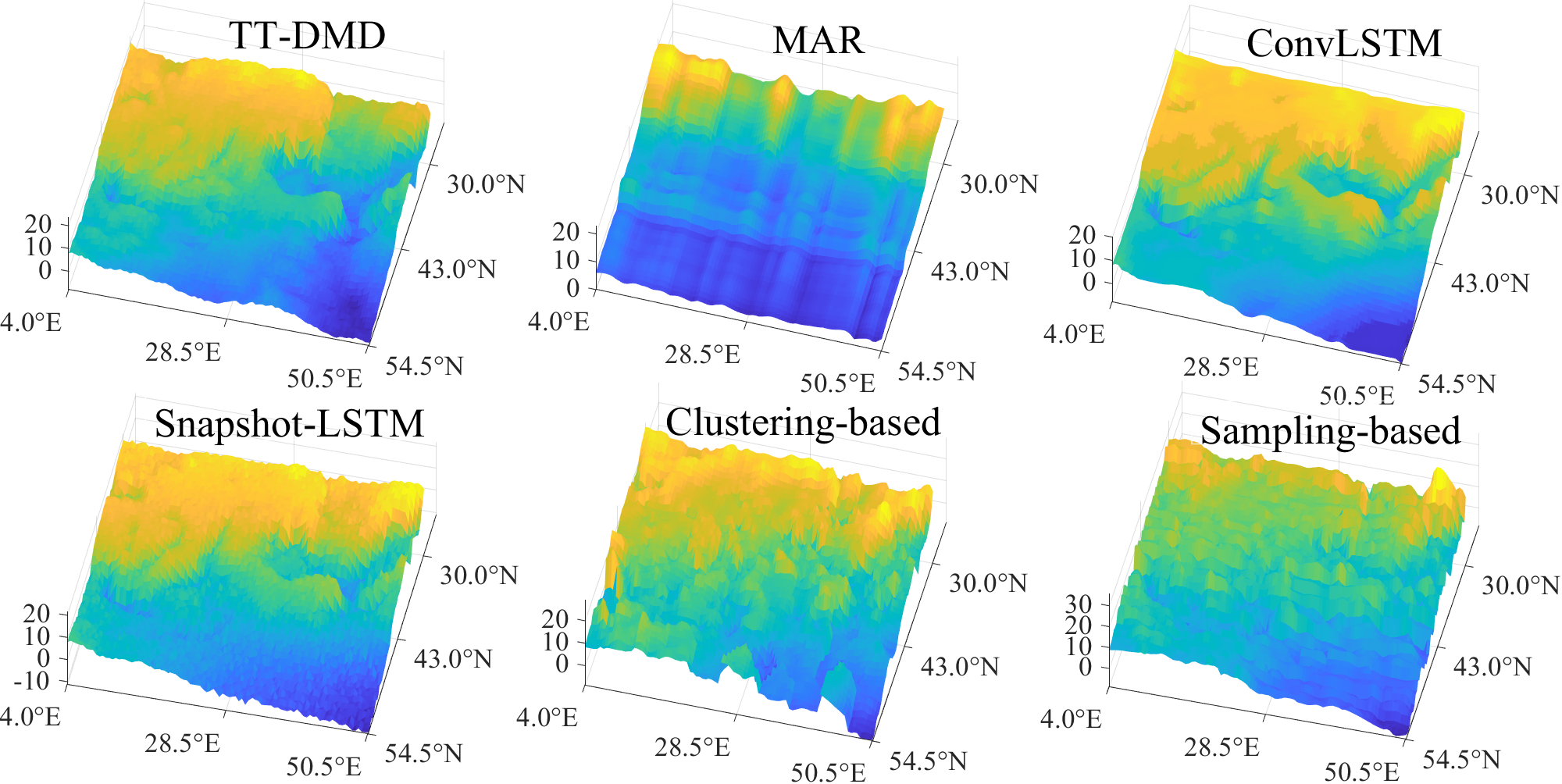}
  \caption{Predicted temperature distributions using different forecasting methods \texttt{2020-08-13}}
  \label{fig:spSSA}
  \vspace{-7mm}
\end{figure}

\begin{figure}[ht]
  \centering
\includegraphics[width=1\textwidth]{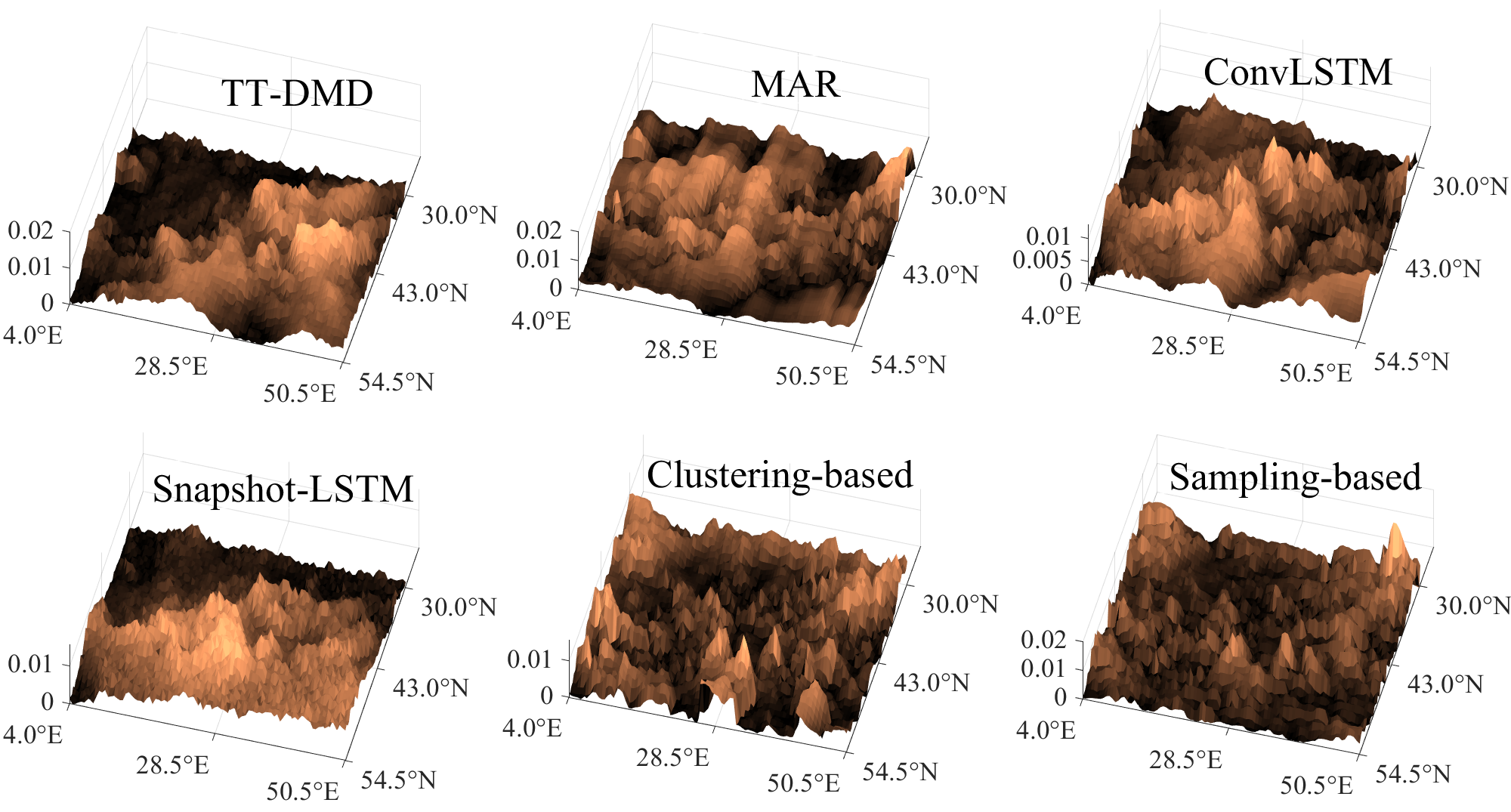}
  \caption{Regional RMSE distribution between predictions and target}
  \label{fig:ErspSSA}
    \vspace{-5mm}
\end{figure}

\noindent \underline{Results:} We compared the 7-steps-ahead predictions of the various spatiotemporal models. In Fig. \ref{fig:tarpred}, we plotted the target temperature field and corresponding predictions by different models across all locations on \texttt{ 2020-08-13} (this is the seventh time step of prediction) in Fig. \ref{fig:spSSA}. We can see that the ConvLSTM and TT-DMD models predict the temperature field quite well. However, the MAR predictions on the seventh time step are the least accurate since this MAR is only of order one. Also, it is interesting to note that models based on dimensionality reduction are pretty good at predicting regional temperature, despite being trained on a small number of selected locations.
\noindent  The RMSE distribution across the locations is shown in Fig. \ref{fig:ErspSSA}, and we infer that the RMSE  for a ConvLSTM is least, while others have nearly similar.  The plots show that error in the sampling-based model is uniformly distributed, whereas error in other models is much higher in some regions.

\begin{table}[t]
\setlength\tabcolsep{4pt}
\caption{Overall Comparison of different spatiotemporal models on short-term prediction (7 step ahead)}
\begin{tabular}{lllllll}
\hline
Metrics & ConvLSTM & TT-DMD & MAR & \begin{tabular}[c]{@{}l@{}}LSTM \\ (Snapshot)\end{tabular} & \begin{tabular}[c]{@{}l@{}}Sampling\\  Lats based\end{tabular} & \begin{tabular}[c]{@{}l@{}}Clustering\\  based\end{tabular} \\ \hline
RMSE & 1.70 & 3.01 & 4.45 & 2.88 & 2.46 & 2.95 \\ \hline
MAE & 1.37 & 2.88 & 3.68 & 2.34 & 1.97 & 2.79 \\ \hline
SMAPE & 22.89 & 45.70 & 52.23 & 49.72 & 30.09 & 33.80 \\ \hline
\begin{tabular}[c]{@{}l@{}}Time \\ (training+\\ inference)\end{tabular} & \begin{tabular}[c]{@{}l@{}}4 mins \\ + 54 sec\end{tabular} & 91 sec & 84 sec & \begin{tabular}[c]{@{}l@{}}4 min 44 sec \\ + 1 min 19 sec\end{tabular} & \begin{tabular}[c]{@{}l@{}}14 mins 2 sec \\ + 7 mins 37 sec\end{tabular} & \begin{tabular}[c]{@{}l@{}}14 mins 49 sec \\ + 8 mins 3 sec\end{tabular} \\ \hline
\end{tabular}
\label{Tab:overallcomp}
  \vspace{-3mm}
\end{table}
\noindent To draw a general conclusion about these models, we compared their performance for seven-step predictions, recorded their training and inference times and other error metrics, and presented the results in Table \ref{Tab:overallcomp}. Based on the results presented in Table \ref{Tab:overallcomp}, we can infer that ConvSTM performed the best while MAR performed the worst in terms of the error metric. The rest show roughly similar behaviour. TT-DMD is optimal in terms of the total time taken for data fitting and predictions. TT-DMD prediction ability is good even though we used the simple variant of DMD; it can be improved significantly, for example, by employing Extended DMD \cite{Williams2014ADA}, Multi-Resolution DMD\cite{kutz2015MDMD} and higher order DMD \cite{le2017HODMD}. Models based on clustering and sampling took significantly longer to train and evaluate, suggesting they are not best suited for long-term predictions. In conclusion, spatiotemporal models that work directly with the original dimensions work significantly better since they capture both spatial and temporal correlations. Furthermore, TT-DMD is a good choice when quick spatiotemporal predictions are required.
\subsubsection{Long-term predictions:}
\label{subsec:Ms_SP}
\noindent We concluded in the preceding section that ConvLSTM, TT-DMD, and MAR models could be easily tested for long-term predictions because they are computationally less expensive. 
This section validates their success for long-term forecasting (250-time steps ahead).\par
\noindent \underline{Training procedure:}~ 
The ConvLSTM training procedure is the same as the short-term prediction training procedure discussed in the previous section, except that the data instances now have a longer lookback (i.e., to incorporate more past states/history) and forecast horizons; we used thirty frames, fifteen for input and fifteen for prediction and used iterative (recursive) prediction strategy to get 250 step predictions, as shown in \cite{recur_mimo1}, \cite{recur_mimo2}. The training for TT-DMD and MAR is the same, but the prediction length has been increased to 250 steps. Although rolling forecasting mechanisms can be used in MAR or TT-DMD, we limited our technique in this work to simple prediction mechanisms only.\par
\begin{figure}[ht]
  \centering
\includegraphics[width=0.98\textwidth ]{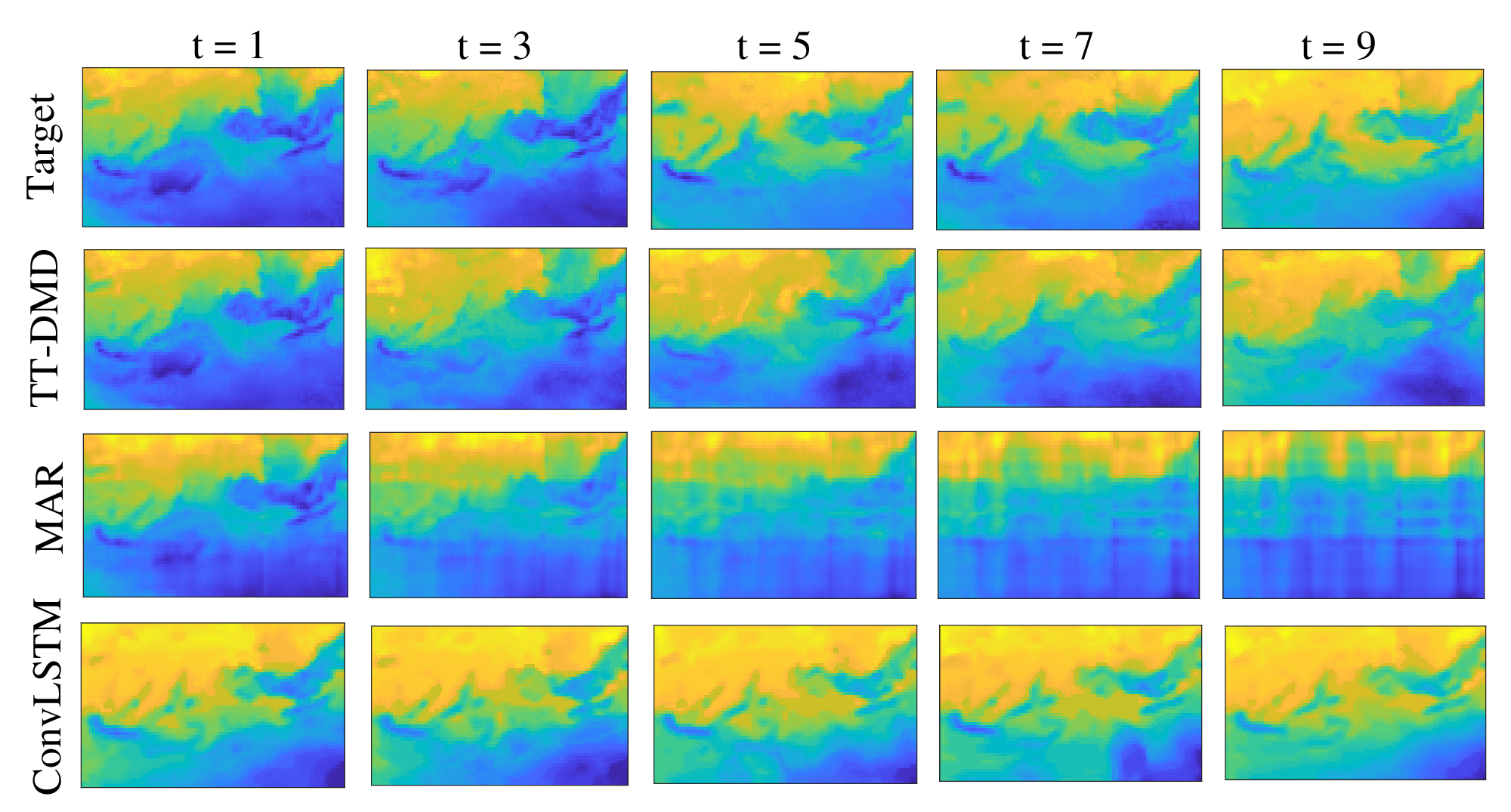}
  \caption{Performance comparison of frame-wise predictions over multiple future time steps.}
  \label{fig:sp_field}
    \vspace{-5mm}
\end{figure}

\noindent \underline{Results:} 
\noindent We compare the predictions of MAR, ConvLSTM, and TT-DMD with the target, as shown in Fig. \ref{fig:sp_field}. From the plots, we can deduce that the MAR model cannot correctly predict multiple steps ahead, whereas TT-DMD and ConvLSTM predictions are quite close to target frames. According to the predictions, TT-DMD is superior to ConvLSTM at capturing spatial information for the first few steps. However, other metrics revealed that ConvLSTM is better when the prediction horizon increases. 

\begin{figure}[h]
  \centering
\includegraphics[width=1\textwidth]{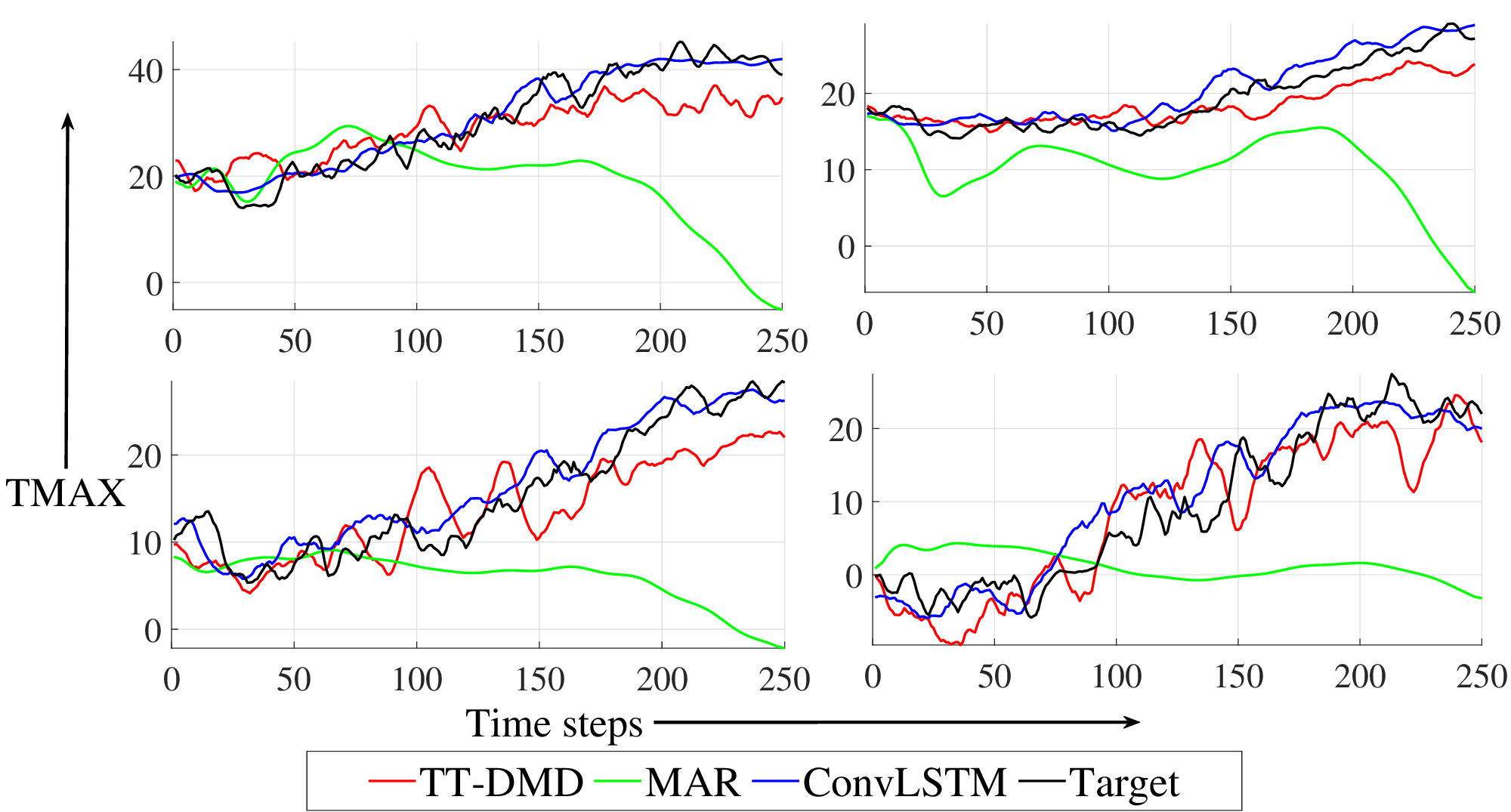}
  \caption{Performance comparison of at randomly selected locations over latitude-longitude grids}
  \label{fig:sp_MSA}
    \vspace{-5mm}
\end{figure}

\noindent Furthermore, we chose four arbitrary locations and compared the performance of different models at each location, i.e., we took four time-series fibers (individual temperature time series) of the spatiotemporal prediction tensor, $\underline{\mathbf{X}}_{predition}$ and visualized them over time. These time series are depicted in Fig. \ref{fig:sp_MSA}. 
\noindent To clearly visualize predictions, the locally weighted scatterplot smoothness (LOWESS) plots of the TT-DMD, ConvLSTM, and MAR predictions are compared. It was found that ConvLSTM and TT-DMD successfully capture trends, and their trajectories are close to the target. In contrast, MAR predictions are unable to follow the target trajectory as MAR does not effectively incorporate long-range space-time correlations.  \ref{Tab:locwise}. Table \ref{Tab:locwise} presents the mean RMSE and SMAPE over thirty distinct locations. For each location, the LOWESS predictions are used to compute the errors against the target prediction. 
\begin{table}[htb]
\caption{The mean RMSE and SMAPE  over thirty distinct locations}
\centering
\begin{tabular}{|c|c|c|c|}
\hline
Metrics & TT-DMD & MAR & ConvLSTM \\ \hline
RMSE & 4.677 & 12.248 & 3.298 \\ \hline
SMAPE & 19.093 & 59.188 & 14.511 \\ \hline
\end{tabular}
\label{Tab:locwise}
\end{table}

\begin{figure}[h]
  \centering
\includegraphics[width=1\textwidth]{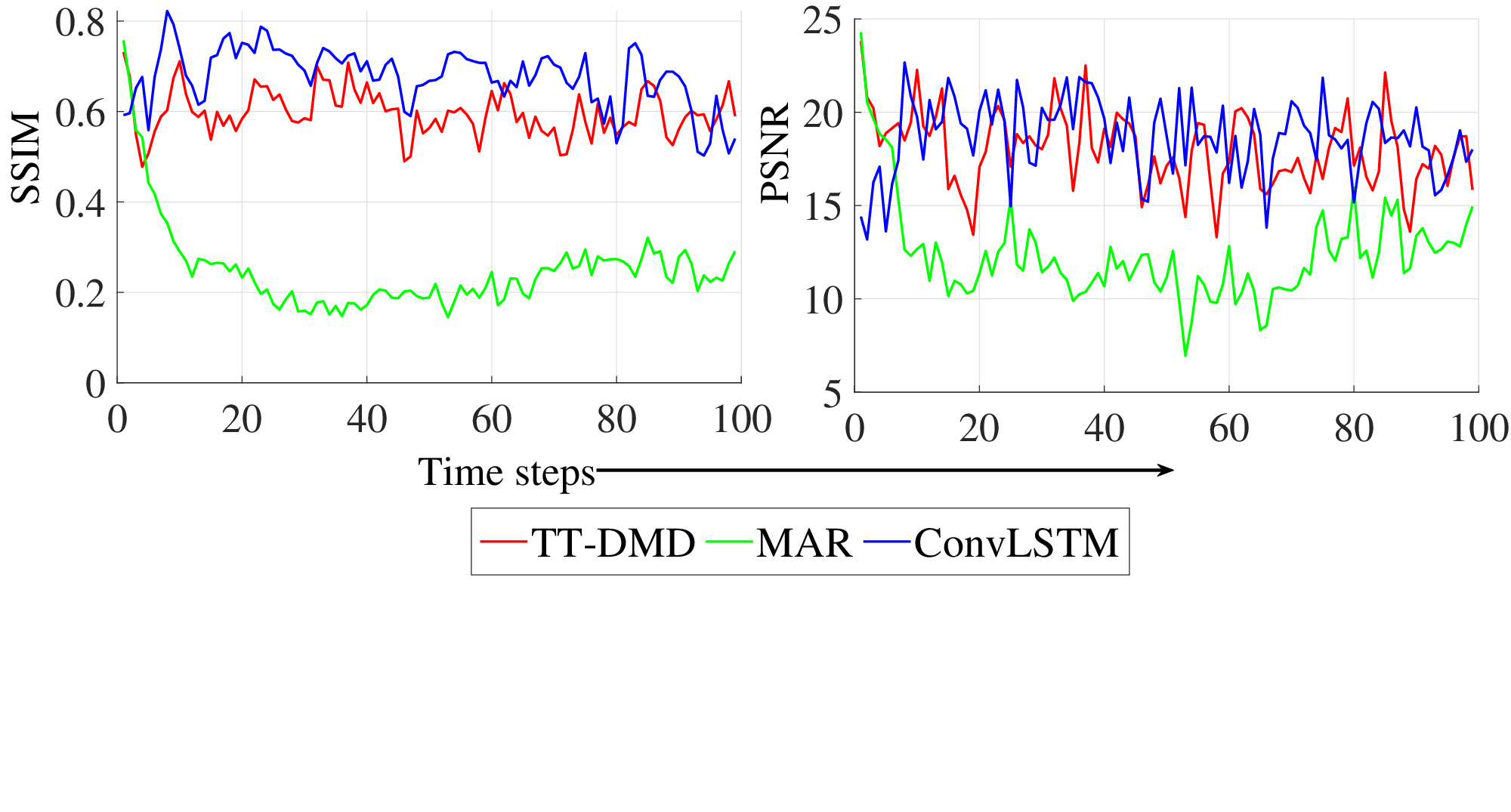}
   \vspace{-20mm}
  \caption{Framewise SSIM and PSNR comparison among different modes.}
  \label{fig:FWError}
\end{figure}

\noindent To assess the spatiotemporal prediction ability among these models, we compare the field distributions over more than 100 prediction steps, frame by frame (slice-wise), with the target. For each time step, we computed the frame-wise mean-square error (MSE), normalised root-mean-square error (NMSE), structural similarity index (SSIM), and peak signal-to-noise ratio (PSNR). Figure \ref{fig:FWError} shows that the SSIM and PSNR of the TT-DMD and ConvLSTM predictions are higher than MAR and remain nearly constant throughout the prediction horizon. This implies that ConvLSTM and TT-DMD provide stable predictions. It's worth noting that despite ConvLSTM being a sophisticated deep-learning model, it only marginally outperforms TT-DMD, which is a simple and interpretable multilinear auto-regression model. 

\noindent In Table \ref{Tab:spMSA}, we listed the mean performance (mean errors/accuracy across all 250 prediction slices) of models on given metrics and discovered that ConvLSTM predictions are 68\% structurally similar to the target, while TT-DMD predictions are 66\% similar. ConvLSTM has the highest average PSNR, followed by TT-DMD. TT-DMD has the lowest mean NRMSE, followed by ConvLSTM. The performance of MAR is the worst among them.
\begin{table}[t]
\caption{Framewise-comparison between MAR, TT-DMD, and ConvLSTM}
\begin{tabular}{|c|c|c|c|}
\hline
Metric & TT-DMD & MAR & ConvLSTM \\ \hline
MSE & 1143.08 & 4327.14 & 986.86 \\ \hline
NRMSE & \textbf{0.22} & 0.46 & 0.23 \\ \hline
\multicolumn{1}{|l|}{PSNR} & 19.13 & 13.74 & 20.83 \\ \hline
\multicolumn{1}{|l|}{SSIM} & 0.66 & 0.24 & 0.68 \\ \hline
\end{tabular}
\label{Tab:spMSA}
\end{table}

\subsubsection*{Additional Experiment}
To further compare with state-of-the-art models, we used the newly developed Python framework for spatiotemporal predictive learning (OpenSTL)\footnote{\url{https://github.com/chengtan9907/OpenSTL}} \cite{tan2023openstl}. The subset of the temperate dataset from the WeatherBench \cite{duben2020weatherbench} was used for experiments.  The subset dataset contains temperature values from 2016 to 2018 with a spatial resolution of 5.625° over a grid of 32 × 64 points. The dataset consists of 17520 (1-hour temporal resolution) temperature distribution maps (32 x 64). The models were trained on 85\% of the data, validated on 7\%, and tested on the remaining samples. In this experiment, we trained the following models: ConvLSTM~\cite{ConvLSTM1}, PhyDNet~\cite{phydnet}, PredRNN~\cite{prednet},  TAU~\cite{tau}, SimVP~\cite{simvp}, MAR~\cite{MAR} and compared with the proposed TT-DMD model. The models from OpenSTL were trained using default parameters. However, the input sequence length (lookback) was changed to 12, and the output sequence length (prediction horizon) was also set to 12. We used the Adam optimizer with a learning rate 1e-3 and batch size equal to 8. Note that the proposed model does not require training, as it is an algebraic method and, therefore, utilizes only standard linear algebra operations to make predictions. The TT-DMD method only takes 51 seconds to reconstruct the data from DMD modes and make predictions. The RMSE and MAE values have been recorded and are summarized in Table~\ref{Tab:addexp}. The sample experiments are available\footnote{\url{https://github.com/ShakirSofi/TensorizingDMD/tree/main/Tensorizing_DMD}}

\begin{table}[h]
\caption{Comparison of different spatiotemporal models (50 step ahead)}
\begin{tabular}{|l|l|l|l|}
\hline
Model    & RMSE & MAE  \\ \hline
ConvLSTM & 2.02 & 5.54 \\ \hline
PhyDNet  & 6.29 & 11.22 \\ \hline
SimVP    & 1.79 & 5.13 \\ \hline
PredRNN  & 2.36 & 6.30 \\ \hline
TAU      & 1.67 & 4.47 \\ \hline
MAR      & 2.86 & 4.08  \\ \hline
TT-DMD   & 2.28 & \textbf{3.25}  \\ \hline
\end{tabular}
\label{Tab:addexp}
\end{table}

The data in this table show that the proposed approach outperforms some sophisticated state-of-the-art models or is comparable in accuracy. This finding also broadly supports the proposed method's practicality. However, these findings cannot be generalized to all types of datasets. We noticed that DL methods perform better when data dynamics change rapidly and exhibit high non-linearity (e.g., extreme weather events). A possible explanation for why the proposed methods outperform some highly sophistical DL models in the experiment above might be that the dynamics in data are not highly non-linear.

To provide a broader picture of these models, we summarized the primary findings in Table~\ref{Tab:Summary}, emphasizing the key advantages and shortcomings of various forecasting models. In this table, we classified the models into four main categories. The first category comprises deep learning-based models, which directly learn spatiotemporal features using 2D/3D convolutions and variants of recurrent neural networks (ConvLSTM, PhyDNet, PredRNN, etc.). The second category includes state-space type models (TT-DMD and MAR), which employ matrix or tensor-based spatiotemporal autoregression. The third category consists of models that squash all spatial features into a vector (snapshot) and then learn the temporal evolution of the snapshots using RNNs, LSTMs, etc. In this category, we do not utilize inter-spatial correlations in modelling. The last category is clustering-based models, which are hybrid; they incorporate local spatial correlations and model temporal evolutions using vector-based models. 

\begin{landscape}
\begin{table}[htb]
\centering
\begin{tabular}{|l|p{7cm}|p{7.2cm}|}
\hline
Model &
  Pros &
  Cons \\ \hline
\begin{tabular}[c]{@{}l@{}}Spatiotemporal DL models\\ (convolutions + RNNs, \\ transformers, etc.)\end{tabular} &
\begin{tabular}[c]{@{}l@{}}1. The prediction accuracy is good.\\ 2. These models effectively capture spatiotemporal \\ correlations.\\ 3. These methods are the preferred choice for\\ predicting extreme weather events.\\ 4. Suitable for long-term forecasting.\end{tabular} &
\begin{tabular}[c]{@{}l@{}}1. A significant amount of training data and time\\ is required to train these models.\\ 2. Compared to traditional time series models,\\ these models may exhibit subpar performance\\ when there are slow varying dynamics over\\ successive time stamps.\end{tabular} \\ \hline
\begin{tabular}[c]{@{}l@{}}Multilinear state-space type\\ models (TT-DMD and MAR)\end{tabular} &
\begin{tabular}[c]{@{}l@{}}1. These models, which are a novel extension of\\ vector autoregression, offer a unique approach\\ to time series analysis and forecasting.\\ 2. These models effectively capture spatiotemporal \\ correlation, especially the TT-DMD\\ uses a tensor framework to incorporate\\ both the spatial and temporal correlations.\\ 3. Provide faster and more accurate forecasts for\\ shorter forecasting horizons.\\ 4. Training and inference time is short for MAR.\\ Interestingly, TT-DMD does not require training\\ since it is an algebraic method. Therefore, it is fast.\\ 5. By utilizing tensor train format TT-DMD breaks\\ curse of dimensionality in comparison other\\ models including MAR.\end{tabular} &
\begin{tabular}[c]{@{}l@{}}1. These models may not work well with the\\ irregularly sampled data (i.e., time series data\\ must be regularly sampled).\\ 2. These models assume that underlying data has\\ some low-rank structure, which may not always\\ be the case.\\ 3. These models may struggle to learn long-term\\ correlations.\\ 4. Optimizing model parameters becomes a challenge\\ on increasing the autoregression order.\end{tabular} \\ \hline
\begin{tabular}[c]{@{}l@{}}Snapshot temporal models\\ (RNNs, transformers, etc.)\end{tabular} &
\begin{tabular}[c]{@{}l@{}}1. Ease of implementation and data preparation.\\ 2. Works well for most prediction problems that\\ do not involve multiple spatial variables.\end{tabular} &
\begin{tabular}[c]{@{}l@{}}1. Prohibits long lookbacks for high-dimensional vectors.\\ 2. Removes inter-spatial correlation by vectorization\\ (in spatiotemporal datasets).\\ 3. Does not perform well for data with high spatial\\ correlations.\end{tabular} \\ \hline
\begin{tabular}[c]{@{}l@{}}Clustering based models\\ (spatial clustering + RNNs, etc.)\end{tabular} &
\begin{tabular}[c]{@{}l@{}}1. In terms of computational cost and the inclusion\\ of spatiotemporal correlations, these models are\\ superior to training separate models for each\\ coordinate point.\\ 2. These models are intuitive and easy to implement.\\ 3. Easy to incorporate domain knowledge, for\\ example, we utilized temperature gradient information\\ to select models for training based on spatial\\ coordinates.\end{tabular} &
\begin{tabular}[c]{@{}l@{}}1. Assumes nearby locations have similar weather\\ patterns, which does not always hold. It may not hold\\ at certain locations, such as shores, river banks,\\ and islands.\\ 2. These models can only capture local spatial\\ correlations and may not be suitable for predicting\\ extreme weather events.\\ 3. These models are not suitable for long-term\\ predictions due to their high training and\\ inference times.\end{tabular} \\ \hline
\end{tabular}

\caption{Summary table}
\label{Tab:Summary}
\end{table}
\end{landscape}
\section{Concluding remarks}
\label{sec:conclusion}
\noindent We discussed various data-driven approaches for spatiotemporal forecasting, and we showed how to condition these models for forecasting weather states over multiple locations simultaneously by exploiting spatiotemporal correlations. We observed that incorporating spatiotemporal correlations improves performance and reduces the computational cost. We proposed an algebraic forecasting method based on tensor-train DMD. We found that its predictions are comparable to those of the state-of-the-art models without the need for training, and it significantly reduces computational time. We evaluated the proposed methods for short-term and long-term forecasting to ensure they can predict temperature fields multiple steps ahead. Moreover, the proposed model is an excellent choice for quick and short-term spatiotemporal predictions. It should be noted that TT-DMD may struggle to learn spatiotemporal correlations from massive, incomplete, or irregularly sampled datasets. Therefore, Future work should take care of these two issues, and it may also be possible to generalize this approach for other variants of DMDs like Extended DMD \cite{Williams2014ADA}, Multi-Resolution DMD\cite{kutz2015MDMD} and higher order DMD \cite{le2017HODMD}.
\section*{Declarations}

\begin{itemize}
\item \textbf{Funding} This research received no external funding.
\item \textbf{Competing interests} The authors affirm that there are no competing interests relevant to the article.
\item \textbf{Availability of data and materials} Details about the dataset can be found in section 5. We have also provided a code to download this dataset (\url{https://github.com/ShakirSofi/MYDatasets})
\item \textbf{Authors' contributions} Not applicable 
\end{itemize}


\bibliography{sn-article}


\begin{thebibliography}{81}
\ifx \bisbn   \undefined \def \bisbn  #1{ISBN #1}\fi
\ifx \binits  \undefined \def \binits#1{#1}\fi
\ifx \bauthor  \undefined \def \bauthor#1{#1}\fi
\ifx \batitle  \undefined \def \batitle#1{#1}\fi
\ifx \bjtitle  \undefined \def \bjtitle#1{#1}\fi
\ifx \bvolume  \undefined \def \bvolume#1{\textbf{#1}}\fi
\ifx \byear  \undefined \def \byear#1{#1}\fi
\ifx \bissue  \undefined \def \bissue#1{#1}\fi
\ifx \bfpage  \undefined \def \bfpage#1{#1}\fi
\ifx \blpage  \undefined \def \blpage #1{#1}\fi
\ifx \burl  \undefined \def \burl#1{\textsf{#1}}\fi
\ifx \doiurl  \undefined \def \doiurl#1{\url{https://doi.org/#1}}\fi
\ifx \betal  \undefined \def \betal{\textit{et al.}}\fi
\ifx \binstitute  \undefined \def \binstitute#1{#1}\fi
\ifx \binstitutionaled  \undefined \def \binstitutionaled#1{#1}\fi
\ifx \bctitle  \undefined \def \bctitle#1{#1}\fi
\ifx \beditor  \undefined \def \beditor#1{#1}\fi
\ifx \bpublisher  \undefined \def \bpublisher#1{#1}\fi
\ifx \bbtitle  \undefined \def \bbtitle#1{#1}\fi
\ifx \bedition  \undefined \def \bedition#1{#1}\fi
\ifx \bseriesno  \undefined \def \bseriesno#1{#1}\fi
\ifx \blocation  \undefined \def \blocation#1{#1}\fi
\ifx \bsertitle  \undefined \def \bsertitle#1{#1}\fi
\ifx \bsnm \undefined \def \bsnm#1{#1}\fi
\ifx \bsuffix \undefined \def \bsuffix#1{#1}\fi
\ifx \bparticle \undefined \def \bparticle#1{#1}\fi
\ifx \barticle \undefined \def \barticle#1{#1}\fi
\bibcommenthead
\ifx \bconfdate \undefined \def \bconfdate #1{#1}\fi
\ifx \botherref \undefined \def \botherref #1{#1}\fi
\ifx \url \undefined \def \url#1{\textsf{#1}}\fi
\ifx \bchapter \undefined \def \bchapter#1{#1}\fi
\ifx \bbook \undefined \def \bbook#1{#1}\fi
\ifx \bcomment \undefined \def \bcomment#1{#1}\fi
\ifx \oauthor \undefined \def \oauthor#1{#1}\fi
\ifx \citeauthoryear \undefined \def \citeauthoryear#1{#1}\fi
\ifx \endbibitem  \undefined \def \endbibitem {}\fi
\ifx \bconflocation  \undefined \def \bconflocation#1{#1}\fi
\ifx \arxivurl  \undefined \def \arxivurl#1{\textsf{#1}}\fi
\csname PreBibitemsHook\endcsname

\bibitem[\protect\citeauthoryear{IRENA}{2020}]{irena}
\begin{botherref}
\oauthor{\bsnm{IRENA}}:
Innovation landscape brief: Advanced forecasting of variable renewable power
  generation, international renewable energy agency, isbn 978-92-9260-179-9.
IRENA,
Abu Dhabi
(2020).
\url{www.irena.org}
\end{botherref}
\endbibitem

\bibitem[\protect\citeauthoryear{Singh and Yassine}{2018}]{singh2018big}
\begin{barticle}
\bauthor{\bsnm{Singh}, \binits{S.}},
\bauthor{\bsnm{Yassine}, \binits{A.}}:
\batitle{Big data mining of energy time series for behavioral analytics and
  energy consumption forecasting}.
\bjtitle{Energies}
\bvolume{11}(\bissue{2}),
\bfpage{452}
(\byear{2018})
\end{barticle}
\endbibitem

\bibitem[\protect\citeauthoryear{Arroyo et~al.}{2011}]{arroyo2011different}
\begin{barticle}
\bauthor{\bsnm{Arroyo}, \binits{J.}},
\bauthor{\bsnm{Esp{\'\i}nola}, \binits{R.}},
\bauthor{\bsnm{Mat{\'e}}, \binits{C.}}:
\batitle{Different approaches to forecast interval time series: a comparison in
  finance}.
\bjtitle{Computational Economics}
\bvolume{37},
\bfpage{169}--\blpage{191}
(\byear{2011})
\end{barticle}
\endbibitem

\bibitem[\protect\citeauthoryear{Jones and Lorenz}{1986}]{jones1986application}
\begin{barticle}
\bauthor{\bsnm{Jones}, \binits{D.}},
\bauthor{\bsnm{Lorenz}, \binits{M.}}:
\batitle{An application of a markov chain noise model to wind generator
  simulation}.
\bjtitle{Mathematics and Computers in Simulation}
\bvolume{28}(\bissue{5}),
\bfpage{391}--\blpage{402}
(\byear{1986})
\end{barticle}
\endbibitem

\bibitem[\protect\citeauthoryear{Bilbao et~al.}{2002}]{bilbao2002air}
\begin{barticle}
\bauthor{\bsnm{Bilbao}, \binits{J.}},
\bauthor{\bsnm{De~Miguel}, \binits{A.H.}},
\bauthor{\bsnm{Kambezidis}, \binits{H.D.}}:
\batitle{Air temperature model evaluation in the north mediterranean belt
  area}.
\bjtitle{Journal of Applied Meteorology}
\bvolume{41}(\bissue{8}),
\bfpage{872}--\blpage{884}
(\byear{2002})
\end{barticle}
\endbibitem

\bibitem[\protect\citeauthoryear{Richardson}{1922}]{nwp1922}
\begin{bbook}
\bauthor{\bsnm{Richardson}, \binits{L.F.}}:
\bbtitle{Weather prediction by numerical process, Cambridge (University Press),
  1922. 4°. Pp. xii + 236. 30s.net},
vol. \bseriesno{48},
pp. \bfpage{282}--\blpage{284}
(\byear{1922}).
\doiurl{10.1002/qj.49704820311}
\end{bbook}
\endbibitem

\bibitem[\protect\citeauthoryear{Charney}{1955}]{nwp_eqn}
\begin{barticle}
\bauthor{\bsnm{Charney}, \binits{J.}}:
\batitle{The use of the primitive equations of motion in numerical prediction}.
\bjtitle{Tellus}
\bvolume{7}(\bissue{1}),
\bfpage{22}--\blpage{26}
(\byear{1955})
\doiurl{10.1111/j.2153-3490.1955.tb01138.x}
\end{barticle}
\endbibitem

\bibitem[\protect\citeauthoryear{Charney et~al.}{1950}]{charney}
\begin{barticle}
\bauthor{\bsnm{Charney}, \binits{J.G.}},
\bauthor{\bsnm{FjÖrtoft}, \binits{R.}},
\bauthor{\bsnm{Neumann}, \binits{J.V.}}:
\batitle{Numerical integration of the barotropic vorticity equation}.
\bjtitle{Tellus}
\bvolume{2}(\bissue{4}),
\bfpage{237}--\blpage{254}
(\byear{1950})
\doiurl{10.3402/tellusa.v2i4.8607}
\end{barticle}
\endbibitem

\bibitem[\protect\citeauthoryear{Kalnay}{2002}]{kalnay}
\begin{bbook}
\bauthor{\bsnm{Kalnay}, \binits{E.}}:
\bbtitle{Atmospheric Modeling, Data Assimilation and Predictability}.
\bpublisher{Cambridge University Press}, \blocation{???}
(\byear{2002}).
\doiurl{10.1017/CBO9780511802270}
\end{bbook}
\endbibitem

\bibitem[\protect\citeauthoryear{Tolstykh and Frolov}{2005}]{issuesNWP}
\begin{barticle}
\bauthor{\bsnm{Tolstykh}, \binits{M.}},
\bauthor{\bsnm{Frolov}, \binits{A.}}:
\batitle{Some current problems in numerical weather prediction}.
\bjtitle{Izvestiya Atmospheric and Oceanic Physics}
\bvolume{41},
\bfpage{285}--\blpage{295}
(\byear{2005})
\end{barticle}
\endbibitem

\bibitem[\protect\citeauthoryear{Evensen}{1994}]{Evensen}
\begin{barticle}
\bauthor{\bsnm{Evensen}, \binits{G.}}:
\batitle{Sequential data assimilation with a nonlinear quasi‐geostrophic
  model using monte carlo methods to forecast error statistics}.
\bjtitle{Journal of Geophysical Research}
\bvolume{99},
\bfpage{10143}--\blpage{10162}
(\byear{1994})
\end{barticle}
\endbibitem

\bibitem[\protect\citeauthoryear{Qin et~al.}{2019}]{qin1}
\begin{barticle}
\bauthor{\bsnm{Qin}, \binits{D.}},
\bauthor{\bsnm{Yu}, \binits{J.}},
\bauthor{\bsnm{Zou}, \binits{G.}},
\bauthor{\bsnm{Yong}, \binits{R.}},
\bauthor{\bsnm{Zhao}, \binits{Q.}},
\bauthor{\bsnm{Zhang}, \binits{B.}}:
\batitle{A novel combined prediction scheme based on cnn and lstm for urban
  pm2.5 concentration}.
\bjtitle{IEEE Access}
\bvolume{7},
\bfpage{20050}--\blpage{20059}
(\byear{2019})
\doiurl{10.1109/ACCESS.2019.2897028}
\end{barticle}
\endbibitem

\bibitem[\protect\citeauthoryear{Shi et~al.}{2015}]{ConvLSTM1}
\begin{bchapter}
\bauthor{\bsnm{Shi}, \binits{X.}},
\bauthor{\bsnm{Chen}, \binits{Z.}},
\bauthor{\bsnm{Wang}, \binits{H.}},
\bauthor{\bsnm{Yeung}, \binits{D.-Y.}},
\bauthor{\bsnm{Wong}, \binits{W.-k.}},
\bauthor{\bsnm{Woo}, \binits{W.-c.}}:
\bctitle{Convolutional lstm network: A machine learning approach for
  precipitation nowcasting}.
In: \bbtitle{Proceedings of the 28th International Conference on Neural
  Information Processing Systems - Volume 1}.
\bsertitle{NIPS'15},
pp. \bfpage{802}--\blpage{810}.
\bpublisher{MIT Press},
\blocation{Cambridge, MA, USA}
(\byear{2015})
\end{bchapter}
\endbibitem

\bibitem[\protect\citeauthoryear{chun Woo}{2014}]{rover}
\begin{bchapter}
\bauthor{\bsnm{Woo}, \binits{W.-c.}}:
\bctitle{Application of optical flow techniques to rainfall nowcasting}.
(\byear{2014})
\end{bchapter}
\endbibitem

\bibitem[\protect\citeauthoryear{Bai et~al.}{2018}]{Bai2018AnEE}
\begin{botherref}
\oauthor{\bsnm{Bai}, \binits{S.}},
\oauthor{\bsnm{Kolter}, \binits{J.Z.}},
\oauthor{\bsnm{Koltun}, \binits{V.}}:
An empirical evaluation of generic convolutional and recurrent networks for
  sequence modeling.
ArXiv
\textbf{abs/1803.01271}
(2018)
\end{botherref}
\endbibitem

\bibitem[\protect\citeauthoryear{Nascimento et~al.}{2021}]{STConvS2S}
\begin{botherref}
\oauthor{\bsnm{Nascimento}, \binits{R.C.}},
\oauthor{\bsnm{Souto}, \binits{Y.M.}},
\oauthor{\bsnm{Ogasawara}, \binits{E.S.}},
\oauthor{\bsnm{Porto}, \binits{F.A.M.}},
\oauthor{\bsnm{Bezerra}, \binits{E.}}:
Stconvs2s: Spatiotemporal convolutional sequence to sequence network for
  weather forecasting.
ArXiv
\textbf{abs/1912.00134}
(2021)
\end{botherref}
\endbibitem

\bibitem[\protect\citeauthoryear{Tran et~al.}{2018}]{Tran2018ACL}
\begin{botherref}
\oauthor{\bsnm{Tran}, \binits{D.}},
\oauthor{\bsnm{Wang}, \binits{H.}},
\oauthor{\bsnm{Torresani}, \binits{L.}},
\oauthor{\bsnm{Ray}, \binits{J.}},
\oauthor{\bsnm{LeCun}, \binits{Y.}},
\oauthor{\bsnm{Paluri}, \binits{M.}}:
A closer look at spatiotemporal convolutions for action recognition.
2018 IEEE/CVF Conference on Computer Vision and Pattern Recognition,
6450--6459
(2018)
\end{botherref}
\endbibitem

\bibitem[\protect\citeauthoryear{Tan et~al.}{2023}]{tan2023openstl}
\begin{bchapter}
\bauthor{\bsnm{Tan}, \binits{C.}},
\bauthor{\bsnm{Li}, \binits{S.}},
\bauthor{\bsnm{Gao}, \binits{Z.}},
\bauthor{\bsnm{Guan}, \binits{W.}},
\bauthor{\bsnm{Wang}, \binits{Z.}},
\bauthor{\bsnm{Liu}, \binits{Z.}},
\bauthor{\bsnm{Wu}, \binits{L.}},
\bauthor{\bsnm{Li}, \binits{S.Z.}}:
\bctitle{Openstl: A comprehensive benchmark of spatio-temporal predictive
  learning}.
In: \bbtitle{Conference on Neural Information Processing Systems Datasets and
  Benchmarks Track}
(\byear{2023})
\end{bchapter}
\endbibitem

\bibitem[\protect\citeauthoryear{Lotter et~al.}{2017}]{prednet}
\begin{botherref}
\oauthor{\bsnm{Lotter}, \binits{W.}},
\oauthor{\bsnm{Kreiman}, \binits{G.}},
\oauthor{\bsnm{Cox}, \binits{D.}}:
Deep predictive coding networks for video prediction and unsupervised learning
(2017)
\end{botherref}
\endbibitem

\bibitem[\protect\citeauthoryear{Le~Guen and Thome}{2020}]{phydnet}
\begin{bchapter}
\bauthor{\bsnm{Le~Guen}, \binits{V.}},
\bauthor{\bsnm{Thome}, \binits{N.}}:
\bctitle{Disentangling physical dynamics from unknown factors for unsupervised
  video prediction}.
In: \bbtitle{2020 IEEE/CVF Conference on Computer Vision and Pattern
  Recognition (CVPR)},
pp. \bfpage{11471}--\blpage{11481}
(\byear{2020})
\end{bchapter}
\endbibitem

\bibitem[\protect\citeauthoryear{Gao et~al.}{2022}]{simvp}
\begin{bchapter}
\bauthor{\bsnm{Gao}, \binits{Z.}},
\bauthor{\bsnm{Tan}, \binits{C.}},
\bauthor{\bsnm{Wu}, \binits{L.}},
\bauthor{\bsnm{Li}, \binits{S.Z.}}:
\bctitle{Simvp: Simpler yet better video prediction}.
In: \bbtitle{2022 IEEE/CVF Conference on Computer Vision and Pattern
  Recognition (CVPR)},
pp. \bfpage{3160}--\blpage{3170}
(\byear{2022})
\end{bchapter}
\endbibitem

\bibitem[\protect\citeauthoryear{Tan et~al.}{2023}]{tau}
\begin{bchapter}
\bauthor{\bsnm{Tan}, \binits{C.}},
\bauthor{\bsnm{Gao}, \binits{Z.}},
\bauthor{\bsnm{Wu}, \binits{L.}},
\bauthor{\bsnm{Xu}, \binits{Y.}},
\bauthor{\bsnm{Xia}, \binits{J.}},
\bauthor{\bsnm{Li}, \binits{S.}},
\bauthor{\bsnm{Li}, \binits{S.Z.}}:
\bctitle{Temporal attention unit: Towards efficient spatiotemporal predictive
  learning}.
In: \bbtitle{2023 IEEE/CVF Conference on Computer Vision and Pattern
  Recognition (CVPR)},
pp. \bfpage{18770}--\blpage{18782}
(\byear{2023})
\end{bchapter}
\endbibitem

\bibitem[\protect\citeauthoryear{Box and Jenkins}{1990}]{box_jenkin}
\begin{bbook}
\bauthor{\bsnm{Box}, \binits{G.E.P.}},
\bauthor{\bsnm{Jenkins}, \binits{G.}}:
\bbtitle{Time Series Analysis, Forecasting and Control}.
\bpublisher{Holden-Day, Inc.},
\blocation{USA}
(\byear{1990})
\end{bbook}
\endbibitem

\bibitem[\protect\citeauthoryear{Delleur and Kavvas}{1978}]{stoch_rain}
\begin{barticle}
\bauthor{\bsnm{Delleur}, \binits{J.}},
\bauthor{\bsnm{Kavvas}, \binits{M.}}:
\batitle{Stochastic models for monthly rainfall forecasting and synthetic
  generation}.
\bjtitle{Journal of Applied Meteorology}
\bvolume{17},
\bfpage{1528}--\blpage{1536}
(\byear{1978})
\doiurl{10.1175/1520-0450(1978)017<1528:SMFMRF>2.0.CO;2}
\end{barticle}
\endbibitem

\bibitem[\protect\citeauthoryear{Zhang et~al.}{2011}]{ssaArima}
\begin{barticle}
\bauthor{\bsnm{Zhang}, \binits{Q.}},
\bauthor{\bsnm{Wang}, \binits{B.-D.}},
\bauthor{\bsnm{He}, \binits{B.}},
\bauthor{\bsnm{Peng}, \binits{Y.}},
\bauthor{\bsnm{Ren}, \binits{M.-L.}}:
\batitle{{Singular Spectrum Analysis and ARIMA Hybrid Model for Annual Runoff
  Forecasting}}.
\bjtitle{Water Resources Management: An International Journal, Published for
  the European Water Resources Association (EWRA)}
\bvolume{25}(\bissue{11}),
\bfpage{2683}--\blpage{2703}
(\byear{2011})
\doiurl{10.1007/s11269-011-9833-y}
\end{barticle}
\endbibitem

\bibitem[\protect\citeauthoryear{Chen et~al.}{2018}]{sarima_nanjing}
\begin{barticle}
\bauthor{\bsnm{Chen}, \binits{P.}},
\bauthor{\bsnm{Niu}, \binits{A.}},
\bauthor{\bsnm{Liu}, \binits{D.}},
\bauthor{\bsnm{Jiang}, \binits{W.}},
\bauthor{\bsnm{Ma}, \binits{B.}}:
\batitle{Time series forecasting of temperatures using sarima: An example from
  nanjing}.
\bjtitle{IOP Conference Series: Materials Science and Engineering}
\bvolume{394},
\bfpage{052024}
(\byear{2018})
\doiurl{10.1088/1757-899X/394/5/052024}
\end{barticle}
\endbibitem

\bibitem[\protect\citeauthoryear{Yu}{2018}]{svmxgb}
\begin{botherref}
\oauthor{\bsnm{Yu}, \binits{X.}}:
Comparison of support vector machine and extreme gradient boosting for
  predicting daily global solar radiation using temperature and precipitation
  in humid subtropical climates: A case study in china.
Energy Conversion and Management
\textbf{164}
(2018)
\doiurl{10.1016/j.enconman.2018.02.087}
\end{botherref}
\endbibitem

\bibitem[\protect\citeauthoryear{Vapnik}{1999}]{svmvapnik}
\begin{bbook}
\bauthor{\bsnm{Vapnik}, \binits{V.}}:
\bbtitle{The Nature of Statistical Learning Theory}.
\bpublisher{Springer}, \blocation{???}
(\byear{1999}).
\doiurl{10.1007/978-1-4757-3264-1}
\end{bbook}
\endbibitem

\bibitem[\protect\citeauthoryear{Chen et~al.}{2015}]{xgboost}
\begin{barticle}
\bauthor{\bsnm{Chen}, \binits{T.}},
\bauthor{\bsnm{He}, \binits{T.}},
\bauthor{\bsnm{Benesty}, \binits{M.}},
\bauthor{\bsnm{Khotilovich}, \binits{V.}},
\bauthor{\bsnm{Tang}, \binits{Y.}},
\bauthor{\bsnm{Cho}, \binits{H.}},
\bauthor{\bsnm{Chen}, \binits{K.}}, \betal:
\batitle{Xgboost: extreme gradient boosting}.
\bjtitle{R package version 0.4-2}
\bvolume{1}(\bissue{4}),
\bfpage{1}--\blpage{4}
(\byear{2015})
\end{barticle}
\endbibitem

\bibitem[\protect\citeauthoryear{Rensheng et~al.}{2004}]{chen1}
\begin{barticle}
\bauthor{\bsnm{Rensheng}, \binits{C.}},
\bauthor{\bsnm{Ersi}, \binits{K.}},
\bauthor{\bsnm{Yang}, \binits{J.}},
\bauthor{\bsnm{Lyu}, \binits{S.}},
\bauthor{\bsnm{Zhao}, \binits{W.}}:
\batitle{Validation of five global radiation models with measured daily data in
  china}.
\bjtitle{Energy Conversion and Management}
\bvolume{45},
\bfpage{1759}--\blpage{1769}
(\byear{2004})
\doiurl{10.1016/j.enconman.2003.09.019}
\end{barticle}
\endbibitem

\bibitem[\protect\citeauthoryear{Scott}{2001}]{scott1}
\begin{barticle}
\bauthor{\bsnm{Scott}, \binits{B.}}:
\batitle{Estimation of solar radiation in australia from rainfall and
  temperature observations}.
\bjtitle{Agricultural and Forest Meteorology}
\bvolume{106},
\bfpage{41}--\blpage{59}
(\byear{2001})
\doiurl{10.1016/S0168-1923(00)00173-8}
\end{barticle}
\endbibitem

\bibitem[\protect\citeauthoryear{Mutlu et~al.}{2008}]{mutlu}
\begin{barticle}
\bauthor{\bsnm{Mutlu}, \binits{E.}},
\bauthor{\bsnm{Chaubey}, \binits{I.}},
\bauthor{\bsnm{Hexmoor}, \binits{H.}},
\bauthor{\bsnm{Bajwa}, \binits{S.G.}}:
\batitle{Comparison of artificial neural network models for hydrologic
  predictions at multiple gauging stations in an agricultural watershed}.
\bjtitle{Hydrological Processes}
\bvolume{22}(\bissue{26}),
\bfpage{5097}--\blpage{5106}
(\byear{2008})
\doiurl{10.1002/hyp.7136}
\end{barticle}
\endbibitem

\bibitem[\protect\citeauthoryear{Qi and Zhang}{2008}]{quzhang}
\begin{barticle}
\bauthor{\bsnm{Qi}, \binits{M.}},
\bauthor{\bsnm{Zhang}, \binits{G.P.}}:
\batitle{Trend time–series modeling and forecasting with neural networks}.
\bjtitle{IEEE Transactions on Neural Networks}
\bvolume{19}(\bissue{5}),
\bfpage{808}--\blpage{816}
(\byear{2008})
\doiurl{10.1109/TNN.2007.912308}
\end{barticle}
\endbibitem

\bibitem[\protect\citeauthoryear{Wei et~al.}{2013}]{hybrid}
\begin{barticle}
\bauthor{\bsnm{Wei}, \binits{S.}},
\bauthor{\bsnm{Yang}, \binits{H.}},
\bauthor{\bsnm{Song}, \binits{J.}},
\bauthor{\bsnm{Abbaspour}, \binits{K.}},
\bauthor{\bsnm{Xu}, \binits{Z.}}:
\batitle{A wavelet-neural network hybrid modelling approach for estimating and
  predicting river monthly flows}.
\bjtitle{Hydrological Sciences Journal}
\bvolume{58}(\bissue{2}),
\bfpage{374}--\blpage{389}
(\byear{2013})
\doiurl{10.1080/02626667.2012.754102}
\end{barticle}
\endbibitem

\bibitem[\protect\citeauthoryear{Anselin}{2013}]{SEMM}
\begin{bbook}
\bauthor{\bsnm{Anselin}, \binits{L.}}:
\bbtitle{Spatial Econometrics: Methods and Models}.
\bsertitle{Studies in Operational Regional Science}.
\bpublisher{Springer}, \blocation{???}
(\byear{2013}).
\burl{https://books.google.ru/books?id=G47tCAAAQBAJ}
\end{bbook}
\endbibitem

\bibitem[\protect\citeauthoryear{Pace et~al.}{1998}]{SEMMapp}
\begin{barticle}
\bauthor{\bsnm{Pace}, \binits{R.}},
\bauthor{\bsnm{Barry}, \binits{R.}},
\bauthor{\bsnm{Clapp}, \binits{J.}},
\bauthor{\bsnm{Rodriquez}, \binits{M.}}:
\batitle{Spatio-temporal autoregressive models of neighborhood effects}.
\bjtitle{The Journal of Real Estate Finance and Economics}
\bvolume{17},
\bfpage{15}--\blpage{33}
(\byear{1998})
\doiurl{10.1023/A:1007799028599}
\end{barticle}
\endbibitem

\bibitem[\protect\citeauthoryear{Shi and Yeung}{2018}]{STML}
\begin{botherref}
\oauthor{\bsnm{Shi}, \binits{X.}},
\oauthor{\bsnm{Yeung}, \binits{D.-Y.}}:
Machine learning for spatiotemporal sequence forecasting: A survey.
ArXiv
\textbf{abs/1808.06865}
(2018)
\end{botherref}
\endbibitem

\bibitem[\protect\citeauthoryear{Lecun et~al.}{1998}]{cnn1}
\begin{barticle}
\bauthor{\bsnm{Lecun}, \binits{Y.}},
\bauthor{\bsnm{Bottou}, \binits{L.}},
\bauthor{\bsnm{Bengio}, \binits{Y.}},
\bauthor{\bsnm{Haffner}, \binits{P.}}:
\batitle{Gradient-based learning applied to document recognition}.
\bjtitle{Proceedings of the IEEE}
\bvolume{86}(\bissue{11}),
\bfpage{2278}--\blpage{2324}
(\byear{1998})
\doiurl{10.1109/5.726791}
\end{barticle}
\endbibitem

\bibitem[\protect\citeauthoryear{Rajagukguk et~al.}{2020}]{en13246623}
\begin{botherref}
\oauthor{\bsnm{Rajagukguk}, \binits{R.A.}},
\oauthor{\bsnm{Ramadhan}, \binits{R.A.A.}},
\oauthor{\bsnm{Lee}, \binits{H.-J.}}:
A review on deep learning models for forecasting time series data of solar
  irradiance and photovoltaic power.
Energies
\textbf{13}(24)
(2020)
\doiurl{10.3390/en13246623}
\end{botherref}
\endbibitem

\bibitem[\protect\citeauthoryear{Pathak et~al.}{2022}]{pathak2022fourcastnet}
\begin{botherref}
\oauthor{\bsnm{Pathak}, \binits{J.}},
\oauthor{\bsnm{Subramanian}, \binits{S.}},
\oauthor{\bsnm{Harrington}, \binits{P.}},
\oauthor{\bsnm{Raja}, \binits{S.}},
\oauthor{\bsnm{Chattopadhyay}, \binits{A.}},
\oauthor{\bsnm{Mardani}, \binits{M.}},
\oauthor{\bsnm{Kurth}, \binits{T.}},
\oauthor{\bsnm{Hall}, \binits{D.}},
\oauthor{\bsnm{Li}, \binits{Z.}},
\oauthor{\bsnm{Azizzadenesheli}, \binits{K.}},
\oauthor{\bsnm{Hassanzadeh}, \binits{P.}},
\oauthor{\bsnm{Kashinath}, \binits{K.}},
\oauthor{\bsnm{Anandkumar}, \binits{A.}}:
FourCastNet: A Global Data-driven High-resolution Weather Model using Adaptive
  Fourier Neural Operators
(2022)
\end{botherref}
\endbibitem

\bibitem[\protect\citeauthoryear{Wu et~al.}{2020}]{deeptf}
\begin{botherref}
\oauthor{\bsnm{Wu}, \binits{N.}},
\oauthor{\bsnm{Green}, \binits{B.}},
\oauthor{\bsnm{Ben}, \binits{X.}},
\oauthor{\bsnm{O'Banion}, \binits{S.}}:
Deep Transformer Models for Time Series Forecasting: The Influenza Prevalence
  Case
(2020)
\end{botherref}
\endbibitem

\bibitem[\protect\citeauthoryear{Lim et~al.}{2020}]{deeptemporal}
\begin{botherref}
\oauthor{\bsnm{Lim}, \binits{B.}},
\oauthor{\bsnm{Arik}, \binits{S.O.}},
\oauthor{\bsnm{Loeff}, \binits{N.}},
\oauthor{\bsnm{Pfister}, \binits{T.}}:
Temporal Fusion Transformers for Interpretable Multi-horizon Time Series
  Forecasting
(2020)
\end{botherref}
\endbibitem

\bibitem[\protect\citeauthoryear{Bi et~al.}{2023}]{ViTforecast}
\begin{botherref}
\oauthor{\bsnm{Bi}, \binits{K.}},
\oauthor{\bsnm{Xie}, \binits{L.}},
\oauthor{\bsnm{Zhang}, \binits{H.}},
\oauthor{\bsnm{Chen}, \binits{X.}},
\oauthor{\bsnm{Gu}, \binits{X.}},
\oauthor{\bsnm{Tian}, \binits{Q.}}:
Accurate medium-range global weather forecasting with 3d neural networks.
Nature,
1--6
(2023)
\doiurl{10.1038/s41586-023-06185-3}
\end{botherref}
\endbibitem

\bibitem[\protect\citeauthoryear{Kolda and Bader}{2009}]{kolda2009tensors}
\begin{barticle}
\bauthor{\bsnm{Kolda}, \binits{T.G.}},
\bauthor{\bsnm{Bader}, \binits{B.W.}}:
\batitle{Tensor decompositions and applications}.
\bjtitle{SIAM Review}
\bvolume{51}(\bissue{3}),
\bfpage{455}--\blpage{500}
(\byear{2009})
\doiurl{10.1137/07070111X}
{\href{https://arxiv.org/abs/https://doi.org/10.1137/07070111X}{{https://doi.org/10.1137/07070111X}}}
\end{barticle}
\endbibitem

\bibitem[\protect\citeauthoryear{Sidiropoulos et~al.}{2017}]{lieven2017tensors}
\begin{barticle}
\bauthor{\bsnm{Sidiropoulos}, \binits{N.D.}},
\bauthor{\bsnm{De~Lathauwer}, \binits{L.}},
\bauthor{\bsnm{Fu}, \binits{X.}},
\bauthor{\bsnm{Huang}, \binits{K.}},
\bauthor{\bsnm{Papalexakis}, \binits{E.E.}},
\bauthor{\bsnm{Faloutsos}, \binits{C.}}:
\batitle{Tensor decomposition for signal processing and machine learning}.
\bjtitle{IEEE Transactions on Signal Processing}
\bvolume{65}(\bissue{13}),
\bfpage{3551}--\blpage{3582}
(\byear{2017})
\doiurl{10.1109/TSP.2017.2690524}
\end{barticle}
\endbibitem

\bibitem[\protect\citeauthoryear{Das and Ghosh}{2020}]{Das_ST}
\begin{barticle}
\bauthor{\bsnm{Das}, \binits{M.}},
\bauthor{\bsnm{Ghosh}, \binits{S.K.}}:
\batitle{Data-driven approaches for spatio-temporal analysis: A survey of the
  state-of-the-arts}.
\bjtitle{J. Comput. Sci. Technol.}
\bvolume{35}(\bissue{3}),
\bfpage{665}--\blpage{696}
(\byear{2020})
\doiurl{10.1007/s11390-020-9349-0}
\end{barticle}
\endbibitem

\bibitem[\protect\citeauthoryear{Wang~K}{2011}]{lat-temp}
\begin{bchapter}
\bauthor{\bsnm{Wang~K}, \binits{C.G.} \bsuffix{Sun~J}}:
\bctitle{Effect of altitude and latitude on surface air temperature across the
  qinghai-tibet plateau, j. mt. sci}.
(\byear{2011}).
\doiurl{10.1007/s11629-011-1090-2}
\end{bchapter}
\endbibitem

\bibitem[\protect\citeauthoryear{Le-Khac et~al.}{2010}]{Dimred}
\begin{bchapter}
\bauthor{\bsnm{Le-Khac}, \binits{N.-A.}},
\bauthor{\bsnm{Bue}, \binits{M.}},
\bauthor{\bsnm{Whelan}, \binits{M.}},
\bauthor{\bsnm{Kechadi}, \binits{M.-T.}}:
\bctitle{A clustering-based data reduction for very large spatio-temporal
  datasets}.
In: \beditor{\bsnm{Cao}, \binits{L.}},
\beditor{\bsnm{Zhong}, \binits{J.}},
\beditor{\bsnm{Feng}, \binits{Y.}} (eds.)
\bbtitle{Advanced Data Mining and Applications},
pp. \bfpage{43}--\blpage{54}.
\bpublisher{Springer},
\blocation{Berlin, Heidelberg}
(\byear{2010})
\end{bchapter}
\endbibitem

\bibitem[\protect\citeauthoryear{Hochreiter and Schmidhuber}{1997}]{lstm_main}
\begin{barticle}
\bauthor{\bsnm{Hochreiter}, \binits{S.}},
\bauthor{\bsnm{Schmidhuber}, \binits{J.}}:
\batitle{{Long Short-Term Memory}}.
\bjtitle{Neural Computation}
\bvolume{9}(\bissue{8}),
\bfpage{1735}--\blpage{1780}
(\byear{1997})
\doiurl{10.1162/neco.1997.9.8.1735}
\end{barticle}
\endbibitem

\bibitem[\protect\citeauthoryear{Chung et~al.}{2014}]{gru}
\begin{botherref}
\oauthor{\bsnm{Chung}, \binits{J.}},
\oauthor{\bsnm{Gulcehre}, \binits{C.}},
\oauthor{\bsnm{Cho}, \binits{K.}},
\oauthor{\bsnm{Bengio}, \binits{Y.}}:
Empirical Evaluation of Gated Recurrent Neural Networks on Sequence Modeling
(2014)
\end{botherref}
\endbibitem

\bibitem[\protect\citeauthoryear{Sutskever et~al.}{2014}]{sequenceLSTM}
\begin{bchapter}
\bauthor{\bsnm{Sutskever}, \binits{I.}},
\bauthor{\bsnm{Vinyals}, \binits{O.}},
\bauthor{\bsnm{Le}, \binits{Q.V.}}:
\bctitle{Sequence to sequence learning with neural networks}.
In: \beditor{\bsnm{Ghahramani}, \binits{Z.}},
\beditor{\bsnm{Welling}, \binits{M.}},
\beditor{\bsnm{Cortes}, \binits{C.}},
\beditor{\bsnm{Lawrence}, \binits{N.D.}},
\beditor{\bsnm{Weinberger}, \binits{K.Q.}} (eds.)
\bbtitle{Advances in Neural Information Processing Systems 27: Annual
  Conference on Neural Information Processing Systems 2014, December 8-13 2014,
  Montreal, Quebec, Canada},
pp. \bfpage{3104}--\blpage{3112}
(\byear{2014})
\end{bchapter}
\endbibitem

\bibitem[\protect\citeauthoryear{Srivastava et~al.}{2015}]{encdecLSTM}
\begin{bchapter}
\bauthor{\bsnm{Srivastava}, \binits{N.}},
\bauthor{\bsnm{Mansimov}, \binits{E.}},
\bauthor{\bsnm{Salakhutdinov}, \binits{R.}}:
\bctitle{Unsupervised learning of video representations using lstms}.
In: \bbtitle{Proceedings of the 32nd International Conference on International
  Conference on Machine Learning - Volume 37}.
\bsertitle{ICML'15},
pp. \bfpage{843}--\blpage{852}.
\bpublisher{JMLR.org}, \blocation{???}
(\byear{2015})
\end{bchapter}
\endbibitem

\bibitem[\protect\citeauthoryear{Chen et~al.}{2021}]{MAR}
\begin{barticle}
\bauthor{\bsnm{Chen}, \binits{R.}},
\bauthor{\bsnm{Xiao}, \binits{H.}},
\bauthor{\bsnm{Yang}, \binits{D.}}:
\batitle{Autoregressive models for matrix-valued time series}.
\bjtitle{Journal of Econometrics}
\bvolume{222}(\bissue{1, Part B}),
\bfpage{539}--\blpage{560}
(\byear{2021})
\doiurl{10.1016/j.jeconom.2020.07.015} .
\bcomment{Annals Issue:Financial Econometrics in the Age of the Digital
  Economy}
\end{barticle}
\endbibitem

\bibitem[\protect\citeauthoryear{SCHMID}{2010}]{schmid}
\begin{barticle}
\bauthor{\bsnm{SCHMID}, \binits{P.J.}}:
\batitle{Dynamic mode decomposition of numerical and experimental data}.
\bjtitle{Journal of Fluid Mechanics}
\bvolume{656},
\bfpage{5}--\blpage{28}
(\byear{2010})
\doiurl{10.1017/S0022112010001217}
\end{barticle}
\endbibitem

\bibitem[\protect\citeauthoryear{Kutz}{2013}]{kutz}
\begin{bbook}
\bauthor{\bsnm{Kutz}, \binits{J.N.}}:
\bbtitle{Data-Driven Modeling and Scientific Computation: Methods for Complex
  Systems and Big Data}.
\bpublisher{Oxford University Press, Inc.},
\blocation{USA}
(\byear{2013})
\end{bbook}
\endbibitem

\bibitem[\protect\citeauthoryear{Penrose}{1956}]{penrose}
\begin{barticle}
\bauthor{\bsnm{Penrose}, \binits{R.}}:
\batitle{On best approximate solutions of linear matrix equations}.
\bjtitle{Mathematical Proceedings of the Cambridge Philosophical Society}
\bvolume{52}(\bissue{1}),
\bfpage{17}--\blpage{19}
(\byear{1956})
\doiurl{10.1017/S0305004100030929}
\end{barticle}
\endbibitem

\bibitem[\protect\citeauthoryear{Klus et~al.}{2018}]{Klus}
\begin{barticle}
\bauthor{\bsnm{Klus}, \binits{S.}},
\bauthor{\bsnm{Gel{\ss}}, \binits{P.}},
\bauthor{\bsnm{Peitz}, \binits{S.}},
\bauthor{\bsnm{Schütte}, \binits{C.}}:
\batitle{Tensor-based dynamic mode decomposition}.
\bjtitle{Nonlinearity}
\bvolume{31}(\bissue{7}),
\bfpage{3359}--\blpage{3380}
(\byear{2018})
\doiurl{10.1088/1361-6544/aabc8f}
\end{barticle}
\endbibitem

\bibitem[\protect\citeauthoryear{Hitchcock}{1927}]{hitchcock1927cpd}
\begin{barticle}
\bauthor{\bsnm{Hitchcock}, \binits{F.L.}}:
\batitle{The expression of a tensor or a polyadic as a sum of products}.
\bjtitle{Journal of Mathematics and Physics}
\bvolume{6}(\bissue{1-4}),
\bfpage{164}--\blpage{189}
(\byear{1927})
\doiurl{10.1002/sapm192761164}
{\href{https://arxiv.org/abs/https://onlinelibrary.wiley.com/doi/pdf/10.1002/sapm192761164}{{https://onlinelibrary.wiley.com/doi/pdf/10.1002/sapm192761164}}}
\end{barticle}
\endbibitem

\bibitem[\protect\citeauthoryear{Harshman}{1970}]{Harshman}
\begin{bchapter}
\bauthor{\bsnm{Harshman}, \binits{R.}}:
\bctitle{Foundations of the parafac procedure: Models and conditions for an
  "explanatory" multi-model factor analysis}.
(\byear{1970})
\end{bchapter}
\endbibitem

\bibitem[\protect\citeauthoryear{De~Lathauwer et~al.}{2000}]{lieven2000mlsvd}
\begin{barticle}
\bauthor{\bsnm{De~Lathauwer}, \binits{L.}},
\bauthor{\bsnm{De~Moor}, \binits{B.}},
\bauthor{\bsnm{Vandewalle}, \binits{J.}}:
\batitle{A multilinear singular value decomposition}.
\bjtitle{SIAM Journal on Matrix Analysis and Applications}
\bvolume{21}(\bissue{4}),
\bfpage{1253}--\blpage{1278}
(\byear{2000})
\doiurl{10.1137/S0895479896305696}
{\href{https://arxiv.org/abs/https://doi.org/10.1137/S0895479896305696}{{https://doi.org/10.1137/S0895479896305696}}}
\end{barticle}
\endbibitem

\bibitem[\protect\citeauthoryear{Tucker}{1966}]{Tucker1966SomeMN}
\begin{barticle}
\bauthor{\bsnm{Tucker}, \binits{L.R.}}:
\batitle{Some mathematical notes on three-mode factor analysis}.
\bjtitle{Psychometrika}
\bvolume{31},
\bfpage{279}--\blpage{311}
(\byear{1966})
\end{barticle}
\endbibitem

\bibitem[\protect\citeauthoryear{Grasedyck}{2010}]{gras2010hT}
\begin{barticle}
\bauthor{\bsnm{Grasedyck}, \binits{L.}}:
\batitle{Hierarchical singular value decomposition of tensors}.
\bjtitle{SIAM Journal on Matrix Analysis and Applications}
\bvolume{31}(\bissue{4}),
\bfpage{2029}--\blpage{2054}
(\byear{2010})
\doiurl{10.1137/090764189}
{\href{https://arxiv.org/abs/https://doi.org/10.1137/090764189}{{https://doi.org/10.1137/090764189}}}
\end{barticle}
\endbibitem

\bibitem[\protect\citeauthoryear{Oseledets}{2011}]{sirIO}
\begin{barticle}
\bauthor{\bsnm{Oseledets}, \binits{I.V.}}:
\batitle{Tensor-train decomposition}.
\bjtitle{SIAM Journal on Scientific Computing}
\bvolume{33}(\bissue{5}),
\bfpage{2295}--\blpage{2317}
(\byear{2011})
\doiurl{10.1137/090752286}
\end{barticle}
\endbibitem

\bibitem[\protect\citeauthoryear{Oseledets et~al.}{2011}]{sirIOpinv}
\begin{botherref}
\oauthor{\bsnm{Oseledets}, \binits{I.}},
\oauthor{\bsnm{Tyrtyshnikov}, \binits{E.}},
\oauthor{\bsnm{Zamarashkin}, \binits{N.}}:
Tensor-train ranks for matrices and their inverses.
Computational Methods in Applied Mathematics
\textbf{11}
(2011)
\doiurl{10.2478/cmam-2011-0022}
\end{botherref}
\endbibitem

\bibitem[\protect\citeauthoryear{Tu et~al.}{2014}]{on_dmd}
\begin{barticle}
\bauthor{\bsnm{Tu}, \binits{J.H.}},
\bauthor{},
\bauthor{\bsnm{Rowley}, \binits{C.W.}},
\bauthor{\bsnm{Luchtenburg}, \binits{D.M.}},
\bauthor{\bsnm{Brunton}, \binits{S.L.}},
\bauthor{\bsnm{and}, \binits{J.N.K.}}:
\batitle{On dynamic mode decomposition: Theory and applications}.
\bjtitle{Journal of Computational Dynamics}
\bvolume{1}(\bissue{2}),
\bfpage{391}--\blpage{421}
(\byear{2014})
\doiurl{10.3934/jcd.2014.1.391}
\end{barticle}
\endbibitem

\bibitem[\protect\citeauthoryear{Sashidhar and
  Kutz}{2022}]{sashidhar2022bagging}
\begin{barticle}
\bauthor{\bsnm{Sashidhar}, \binits{D.}},
\bauthor{\bsnm{Kutz}, \binits{J.N.}}:
\batitle{Bagging, optimized dynamic mode decomposition for robust, stable
  forecasting with spatial and temporal uncertainty quantification}.
\bjtitle{Philosophical Transactions of the Royal Society A}
\bvolume{380}(\bissue{2229}),
\bfpage{20210199}
(\byear{2022})
\end{barticle}
\endbibitem

\bibitem[\protect\citeauthoryear{Dylewsky et~al.}{2022}]{dylewsky2022sfdmd}
\begin{barticle}
\bauthor{\bsnm{Dylewsky}, \binits{D.}},
\bauthor{\bsnm{Barajas-Solano}, \binits{D.}},
\bauthor{\bsnm{Ma}, \binits{T.}},
\bauthor{\bsnm{Tartakovsky}, \binits{A.M.}},
\bauthor{\bsnm{Kutz}, \binits{J.N.}}:
\batitle{Stochastically forced ensemble dynamic mode decomposition for
  forecasting and analysis of near-periodic systems}.
\bjtitle{IEEE Access}
\bvolume{10},
\bfpage{33440}--\blpage{33448}
(\byear{2022})
\doiurl{10.1109/ACCESS.2022.3161438}
\end{barticle}
\endbibitem

\bibitem[\protect\citeauthoryear{Yue et~al.}{2022}]{yue2022pdmd}
\begin{barticle}
\bauthor{\bsnm{Yue}, \binits{W.}},
\bauthor{\bsnm{Liu}, \binits{Q.}},
\bauthor{\bsnm{Ruan}, \binits{Y.}},
\bauthor{\bsnm{Qian}, \binits{F.}},
\bauthor{\bsnm{Meng}, \binits{H.}}:
\batitle{A prediction approach with mode decomposition-recombination technique
  for short-term load forecasting}.
\bjtitle{Sustainable Cities and Society}
\bvolume{85},
\bfpage{104034}
(\byear{2022})
\doiurl{10.1016/j.scs.2022.104034}
\end{barticle}
\endbibitem

\bibitem[\protect\citeauthoryear{Liew et~al.}{2022}]{liew2022streaming}
\begin{barticle}
\bauthor{\bsnm{Liew}, \binits{J.}},
\bauthor{\bsnm{G{\"o}{\c{c}}men}, \binits{T.}},
\bauthor{\bsnm{Lio}, \binits{W.H.}},
\bauthor{\bsnm{Larsen}, \binits{G.C.}}:
\batitle{Streaming dynamic mode decomposition for short-term forecasting in
  wind farms}.
\bjtitle{Wind Energy}
\bvolume{25}(\bissue{4}),
\bfpage{719}--\blpage{734}
(\byear{2022})
\end{barticle}
\endbibitem

\bibitem[\protect\citeauthoryear{Cheng et~al.}{2022}]{cheng2022real}
\begin{barticle}
\bauthor{\bsnm{Cheng}, \binits{Z.}},
\bauthor{\bsnm{Trepanier}, \binits{M.}},
\bauthor{\bsnm{Sun}, \binits{L.}}:
\batitle{Real-time forecasting of metro origin-destination matrices with
  high-order weighted dynamic mode decomposition}.
\bjtitle{Transportation science}
\bvolume{56}(\bissue{4}),
\bfpage{904}--\blpage{918}
(\byear{2022})
\end{barticle}
\endbibitem

\bibitem[\protect\citeauthoryear{Mansouri et~al.}{2023}]{mansouri2023weather}
\begin{barticle}
\bauthor{\bsnm{Mansouri}, \binits{A.}},
\bauthor{\bsnm{Abolmasoumi}, \binits{A.H.}},
\bauthor{\bsnm{Ghadimi}, \binits{A.A.}}:
\batitle{Weather sensitive short term load forecasting using dynamic mode
  decomposition with control}.
\bjtitle{Electric Power Systems Research}
\bvolume{221},
\bfpage{109387}
(\byear{2023})
\end{barticle}
\endbibitem

\bibitem[\protect\citeauthoryear{Filho and Lopes~dos Santos}{2019}]{mch_DMD}
\begin{barticle}
\bauthor{\bsnm{Filho}, \binits{E.V.}},
\bauthor{\bsnm{Santos}, \binits{P.}}:
\batitle{A dynamic mode decomposition approach with hankel blocks to forecast
  multi-channel temporal series}.
\bjtitle{IEEE Control Systems Letters}
\bvolume{3}(\bissue{3}),
\bfpage{739}--\blpage{744}
(\byear{2019})
\doiurl{10.1109/LCSYS.2019.2917811}
\end{barticle}
\endbibitem

\bibitem[\protect\citeauthoryear{Botchkarev}{2018}]{perfmet}
\begin{botherref}
\oauthor{\bsnm{Botchkarev}, \binits{A.}}:
Performance metrics (error measures) in machine learning regression,
  forecasting and prognostics: Properties and typology.
ArXiv
\textbf{abs/1809.03006}
(2018)
\end{botherref}
\endbibitem

\bibitem[\protect\citeauthoryear{Imaizumi and
  Hayashi}{2017}]{pmlr-v70-imaizumi17a}
\begin{bchapter}
\bauthor{\bsnm{Imaizumi}, \binits{M.}},
\bauthor{\bsnm{Hayashi}, \binits{K.}}:
\bctitle{Tensor decomposition with smoothness}.
In: \beditor{\bsnm{Precup}, \binits{D.}},
\beditor{\bsnm{Teh}, \binits{Y.W.}} (eds.)
\bbtitle{Proceedings of the 34th International Conference on Machine Learning}.
\bsertitle{Proceedings of Machine Learning Research},
vol. \bseriesno{70},
pp. \bfpage{1597}--\blpage{1606}.
\bpublisher{PMLR}, \blocation{???}
(\byear{2017}).
\burl{https://proceedings.mlr.press/v70/imaizumi17a.html}
\end{bchapter}
\endbibitem

\bibitem[\protect\citeauthoryear{Yokota et~al.}{2016}]{yakota2016smooth}
\begin{barticle}
\bauthor{\bsnm{Yokota}, \binits{T.}},
\bauthor{\bsnm{Zhao}, \binits{Q.}},
\bauthor{\bsnm{Cichocki}, \binits{A.}}:
\batitle{Smooth parafac decomposition for tensor completion}.
\bjtitle{IEEE Transactions on Signal Processing}
\bvolume{64}(\bissue{20}),
\bfpage{5423}--\blpage{5436}
(\byear{2016})
\doiurl{10.1109/TSP.2016.2586759}
\end{barticle}
\endbibitem

\bibitem[\protect\citeauthoryear{Williams et~al.}{2014}]{Williams2014ADA}
\begin{barticle}
\bauthor{\bsnm{Williams}, \binits{M.O.}},
\bauthor{\bsnm{Kevrekidis}, \binits{I.G.}},
\bauthor{\bsnm{Rowley}, \binits{C.W.}}:
\batitle{A data–driven approximation of the koopman operator: Extending
  dynamic mode decomposition}.
\bjtitle{Journal of Nonlinear Science}
\bvolume{25},
\bfpage{1307}--\blpage{1346}
(\byear{2014})
\end{barticle}
\endbibitem

\bibitem[\protect\citeauthoryear{Kutz et~al.}{2015}]{kutz2015MDMD}
\begin{bchapter}
\bauthor{\bsnm{Kutz}, \binits{J.}},
\bauthor{\bsnm{Fu}, \binits{X.}},
\bauthor{\bsnm{Brunton}, \binits{S.}},
\bauthor{\bsnm{Erichson}, \binits{N.}}:
\bctitle{Multi-resolution dynamic mode decomposition for foreground/background
  separation and object tracking},
pp. \bfpage{921}--\blpage{929}
(\byear{2015}).
\doiurl{10.1109/ICCVW.2015.122}
\end{bchapter}
\endbibitem

\bibitem[\protect\citeauthoryear{Le~Clainche and Vega}{2017}]{le2017HODMD}
\begin{barticle}
\bauthor{\bsnm{Le~Clainche}, \binits{S.}},
\bauthor{\bsnm{Vega}, \binits{J.M.}}:
\batitle{Higher order dynamic mode decomposition}.
\bjtitle{SIAM Journal on Applied Dynamical Systems}
\bvolume{16}(\bissue{2}),
\bfpage{882}--\blpage{925}
(\byear{2017})
\doiurl{10.1137/15M1054924}
{\href{https://arxiv.org/abs/https://doi.org/10.1137/15M1054924}{{https://doi.org/10.1137/15M1054924}}}
\end{barticle}
\endbibitem

\bibitem[\protect\citeauthoryear{An and Anh}{2015}]{recur_mimo1}
\begin{bchapter}
\bauthor{\bsnm{An}, \binits{N.H.}},
\bauthor{\bsnm{Anh}, \binits{D.T.}}:
\bctitle{Comparison of strategies for multi-step-ahead prediction of time
  series using neural network}.
In: \bbtitle{2015 International Conference on Advanced Computing and
  Applications (ACOMP)},
pp. \bfpage{142}--\blpage{149}
(\byear{2015}).
\doiurl{10.1109/ACOMP.2015.24}
\end{bchapter}
\endbibitem

\bibitem[\protect\citeauthoryear{Chandra et~al.}{2021}]{recur_mimo2}
\begin{barticle}
\bauthor{\bsnm{Chandra}, \binits{R.}},
\bauthor{\bsnm{Goyal}, \binits{S.}},
\bauthor{\bsnm{Gupta}, \binits{R.}}:
\batitle{Evaluation of deep learning models for multi-step ahead time series
  prediction}.
\bjtitle{IEEE Access}
\bvolume{9},
\bfpage{83105}--\blpage{83123}
(\byear{2021})
\doiurl{10.1109/ACCESS.2021.3085085}
\end{barticle}
\endbibitem

\bibitem[\protect\citeauthoryear{Rasp et~al.}{2020}]{duben2020weatherbench}
\begin{botherref}
\oauthor{\bsnm{Rasp}, \binits{S.}},
\oauthor{\bsnm{Dueben}, \binits{P.D.}},
\oauthor{\bsnm{Scher}, \binits{S.}},
\oauthor{\bsnm{Weyn}, \binits{J.A.}},
\oauthor{\bsnm{Mouatadid}, \binits{S.}},
\oauthor{\bsnm{Thuerey}, \binits{N.}}:
Weatherbench: A benchmark data set for data-driven weather forecasting.
Journal of Advances in Modeling Earth Systems
\textbf{12}(11)
(2020)
\doiurl{10.1029/2020MS002203}
\end{botherref}
\endbibitem

\end{thebibliography}

\end{document}